\documentclass[10pt]{IEEEtran}
\ifCLASSOPTIONcompsoc
  \usepackage[nocompress]{cite}
\else
  \usepackage{cite}
\fi

\usepackage{amsmath}
\usepackage{amssymb}
\usepackage{amsthm}
\usepackage{hyperref}
\usepackage{multirow}
\usepackage{graphicx}
\usepackage{adjustbox}
\usepackage[table]{xcolor}
\usepackage{booktabs,tabularx}
\usepackage{microtype}
\usepackage{float}
\usepackage{enumitem}

\DeclareMathOperator*{\argmax}{arg\,max}
\DeclareMathOperator*{\argmin}{arg\,min}

\begin{document}
\title{A Coreset Selection of Coreset Selection Literature: Introduction and Recent Advances}
\author{Brian B. Moser$^{1,2}$, Arundhati S. Shanbhag$^{1,2}$, Stanislav Frolov$^{1}$, \\Federico Raue$^{1}$, Joachim Folz$^{1}$, Andreas Dengel$^{1,2}$\\
$^1$ German Research Center for Artificial Intelligence (DFKI), Germany\\
$^2$ RPTU Kaiserslautern-Landau, Germany\\
{\tt\small first.second@dfki.de}}
\IEEEtitleabstractindextext{%
\begin{abstract}
Coreset selection targets the challenge of finding a small, representative subset of a large dataset that preserves essential patterns for effective machine learning. 
Although several surveys have examined data reduction strategies before, most focus narrowly on either classical geometry-based methods or active learning techniques. 
In contrast, this survey presents a more comprehensive view by unifying three major lines of coreset research, namely, training-free, training-oriented, and label-free approaches, into a single taxonomy. 
We present subfields often overlooked by existing work, including submodular formulations, bilevel optimization, and recent progress in pseudo-labeling for unlabeled datasets. 
Additionally, we examine how pruning strategies influence generalization and neural scaling laws, offering new insights that are absent from prior reviews. 
Finally, we compare these methods under varying computational, robustness, and performance demands and highlight open challenges, such as robustness, outlier filtering, and adapting coreset selection to foundation models, for future research.
\end{abstract}

\begin{IEEEkeywords}
Computer Science, Artificial Intelligence, Coreset Selection, Deep Learning, Survey.
\end{IEEEkeywords}}

\maketitle

\IEEEdisplaynontitleabstractindextext

\IEEEpeerreviewmaketitle

\ifCLASSOPTIONcompsoc
\IEEEraisesectionheading{\section{Introduction}\label{sec:introduction}}
\else
\section{Introduction}
\label{sec:introduction}

\fi
\IEEEPARstart{A}{t} the heart of coreset selection lies a fundamental research question: How can we identify a small yet representative subset of a massive dataset that preserves the critical structure or distribution needed for accurate learning?
This problem was originally proposed by \emph{Agarwal et al.} \cite{agarwal2005geometric} to compute the smallest $k$-ball cover for a set of points.
In other words, the parameter $k$ is the \emph{number} of balls that we are allowed to use, and the goal is to minimize how large those balls must be to cover all points.
Since then, coreset research has been extended to a wide range of covering problems with profound implications for deep learning \cite{karnin2019discrepancy, bhalerao2024fine, carmel2025coresets, tereshchenko2024coreset}.

This is particularly relevant in the current era of large-scale datasets and resource-intensive deep learning models \cite{ramesh2022hierarchical, bengio2019extreme}. 
Training on millions of images or text snippets can be prohibitively expensive, both in terms of computation time and memory footprint, as well as energy costs and carbon emissions associated with extensive GPU usage \cite{wang2018dataset, csiba2018importance, zheng2022coverage}. 
By focusing on a representative coreset, practitioners can reduce training time, maintain strong performance, and reduce carbon footprints~\cite{katharopoulos2018not, bhalerao2024fine}.
In real-world scenarios, such as autonomous driving, edge computing, and privacy-constrained applications, strict limits on storage and latency make data-reduction strategies crucial for viable deployment~\cite{ganguli2022predictability, yang2024data, bhore2025fast}.

Despite early foundational work and several surveys and benchmarks, the field now spans a broad range of coreset definitions and reduction objectives that differ substantially in formulation, assumptions, and the signals they exploit \cite{agarwal2005geometric, guo2022deepcore, feldman2020core, settles2009active}. 
This divergence in definitions and methodologies highlights the evolution of the field and motivates the need for a unifying survey. 
As a result, existing reviews often cover only a slice of the landscape (e.g., geometric coresets or active learning), leaving it unclear how methods relate, where they overlap, and which assumptions drive their behavior.
In particular, the same method can be described under different problem statements (coverage, informativeness, optimization).

To address this fragmentation, we introduce a taxonomy that unifies training-free, training-oriented, and label-free coreset selection under one view. 
Using this structure, we survey classical baselines (random, geometry-based), training-driven selectors (loss/gradient/boundary, submodular, gradient matching, bilevel), and emerging label-free approaches (clustering, foundation-model embeddings).
Beyond algorithmic insights, we analyze the theoretical foundations of coreset selection and investigate how pruning strategies affect generalization, neural scaling laws, and dataset redundancy. 
Empirical comparisons highlight the trade-offs between computational efficiency and model performance across different methods.

\autoref{sec:definitions} lays out definitions of coreset selection and active learning, a closely related field.
The connection between these two areas is crucial because many coreset selection strategies naturally extend to active learning by selecting which samples to label.
\autoref{sec:training-free} presents training-free selection techniques, such as random sampling and geometry-based methods that still provide a strong baseline for more sophisticated methods. 
\autoref{sec:training-oriented} covers all approaches that fall under the umbrella term of training-oriented methods. They exploit information about training dynamics on the full dataset, such as scoring metrics, decision boundary estimations, submodular functions, gradient matching, and bilevel optimization.
Next, \autoref{sec:blind} complements the previous sections by examining label-free coreset selection with pseudo-labeling approaches or vision-language models.
To provide practitioners with additional guidelines, \autoref{sec:theoretical-remarks} provides more theoretical insights about coreset selection and when to apply which method.
\autoref{sec:applications} gives an overview of coreset applications to various fields in machine learning. 
Finally, this work summarizes and points to future directions in \autoref{sec:discussion} and \autoref{sec:conclusion}, respectively.

\newpage

\begin{figure}[t!]
    \begin{center}
        \includegraphics[width=.75\columnwidth]{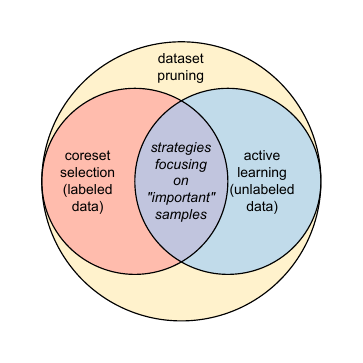}
        \caption{\label{fig:venn}
        Venn diagram of coreset selection and active learning. Coreset selection typically assumes a labeled dataset, whereas active learning uses small labeled datasets and a large label-free pool. Many techniques are shared: both coreset selection and active learning rely on criteria (e.g., uncertainty, diversity, and coverage) to select data subsets.
        }
    \end{center}
\end{figure}

\section{Setting and Terminology}
\label{sec:definitions}
In this section, we formalize the coreset selection problem and highlight its close relationship to active learning, as shown in \autoref{fig:venn}. In \autoref{tab:notation}, we list all typical notations used throughout the survey. We also briefly touch upon theoretical perspectives that will guide deeper insights in later sections. 
To avoid ambiguity across closely related data-reduction settings, \autoref{tab:boundaries} summarizes the key distinctions between coreset selection, active learning, and label-free variants in terms of supervision, outputs, and available training signals.

\subsection{Problem Definition: Coreset Selection}
\label{sec:probdef}
We begin with the classic and standard image classification setup, where the training dataset $\mathcal{T} = \{(\mathbf{x}_i, y_i)\}_{i=1}^{N}$ consists of $N$ i.i.d. samples drawn from an underlying data distribution $P$. 
We define $\mathbf{x}_i\in\mathcal X$ as the input, and $y_i\in\mathcal Y$ as its corresponding ground-truth label.
Coreset selection aims to derive a subset $\mathcal{S} \subset \mathcal{T}$ with \(|\mathcal{S}| \ll | \mathcal{T}|\) such that a model $\theta^\mathcal{S}$ trained on $\mathcal{S}$ generalizes as well as a model $\theta^\mathcal{T}$ trained on the full dataset $\mathcal{T}$:
\begin{equation}
  \mathcal{S}^* = \mathop{\arg\min}_{\mathcal{S} \subset \mathcal{T}: \frac{|\mathcal{S}|}{|\mathcal{T}|} \leq 1 - \alpha} \mathbb{E}_{\mathbf{x},y \sim P}[\mathcal{L}(\mathbf{x},y; \theta^\mathcal{S})],
\end{equation}
where $\alpha \in (0, 1)$ is the pruning ratio and $\mathcal{L}$ is a loss function (usually cross-entropy).

\begin{figure}[t!]
    \begin{center}
        \includegraphics[width=.7\columnwidth]{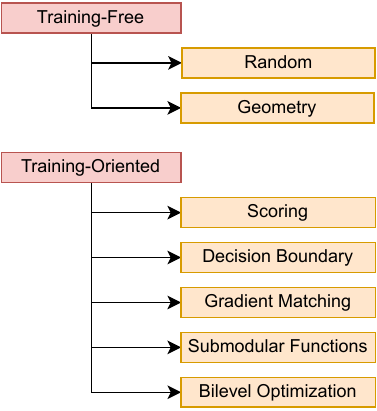}
        \caption{\label{fig:taxonomy}
        Taxonomy used in this survey. We differentiate between training-free and training-oriented methods.
        }
    \end{center}
\end{figure}

Although the above definition is straightforward in principle, \emph{finding} a suitable $\mathcal{S}$ is challenging \cite{agarwal2005geometric,feldman2020core, bachem2017practical}. 
It requires deciding which criteria best measure ``importance'' or ``representativeness'' for a given sample $\mathbf{x}$.
To put it differently, a quantitative way to enforce an order over the input samples $\mathbf{x}$ is needed to guide the selection process \cite{nogueira2018stability, song2022adaptive, xiao2025rethinking}. 
In the following sections, we explore two high-level paradigms and use the taxonomy shown in \autoref{fig:taxonomy}.
\begin{itemize}
    \item \textbf{Training-Free:} Methods that use geometric or statistical properties (e.g., distance in feature space or diversity measures) from the samples alone to select $\mathcal{S}$.
    \item \textbf{Training-Oriented:} Methods that exploit partial or full training signals (e.g., loss values, gradients, forgetting events) to identify the most informative samples for the specific model in use.
\end{itemize}

Most coreset works implicitly assume access to labels, either because they enforce \emph{class-wise} budgets (for class balance) or because the selection directly exploits supervised signals such as per-sample loss, margins, or gradients.
When labels are not used during selection, the setting is typically referred to as \emph{label-free} (also called \emph{unsupervised}/ \emph{blind}): the subset is chosen purely from the inputs.
This distinction is particularly important in scenarios where labels are expensive or unavailable.

\begin{table}[t!]
\centering
\caption{Notation used throughout the survey.}
\label{tab:notation}
\renewcommand{\arraystretch}{1.10}
\setlength{\tabcolsep}{5pt}
\footnotesize
\begin{tabularx}{\columnwidth}{@{} p{0.28\columnwidth} X @{}}
\toprule
\textbf{Symbol} & \textbf{Meaning} \\
\midrule
$\mathcal{T}=\{(\mathbf{x}_i,y_i)\}_{i=1}^{N}$ & Full labeled training dataset with $N$ samples. \\
$\mathcal{S}\subset \mathcal{T}$ & Selected coreset (subset of training data). \\
$|\mathcal{S}|/|\mathcal{T}| \le 1-\alpha$ & Budget constraint; $\alpha\in(0,1)$ is the pruning ratio. \\
$\theta^{\mathcal{S}},\; \theta^{\mathcal{T}}$ & Model parameters trained on $\mathcal{S}$ vs.\ on full $\mathcal{T}$. \\
$\mathcal{T}_{\mathcal{U}},\; \mathcal{T}_{\mathcal{L}}$ & Label-free pool and labeled set in active learning / semi-supervised settings. \\
$\mathcal{Q}\subset \mathcal{T}_{\mathcal{U}}$ & Query batch selected for annotation in active learning. \\
$B_d(\mathbf{x},r)$ & Ball of radius $r$ centered at $\mathbf{x}$ under metric $d$. \\
$f:2^V\rightarrow \mathbb{R}$ & Submodular set function defined on ground set $V$; optimized via greedy maximization. \\
$\chi_t(\mathbf{x},y),\; \chi_t^*(\mathbf{x},y)$ & GraNd score (gradient norm) and EL2N approximation at epoch $t$. \\
$s(\mathbf{x},y)$ & Sample sensitivity; worst-case relative loss contribution. \\
$\gamma_j$ & Sample weight / step-size coefficient in gradient-matching methods (CRAIG/GradMatch). \\
$\lambda$ & Regularization / trade-off scalar (e.g., in GradMatch/RETRIEVE objectives). \\
$\epsilon,\; \beta,\; \beta'$ & Tolerance threshold; scaling-law exponents for $\epsilon(\mathcal{T})\propto|\mathcal{T}|^{-\beta}$ and pruned $\epsilon(\mathcal{S})\propto|\mathcal{S}|^{-\beta'}$. \\
\bottomrule
\end{tabularx}
\end{table}

\subsection{Problem Definition: Active Learning}
\begin{figure}[t!]
    \begin{center}
        \includegraphics[width=.856\columnwidth]{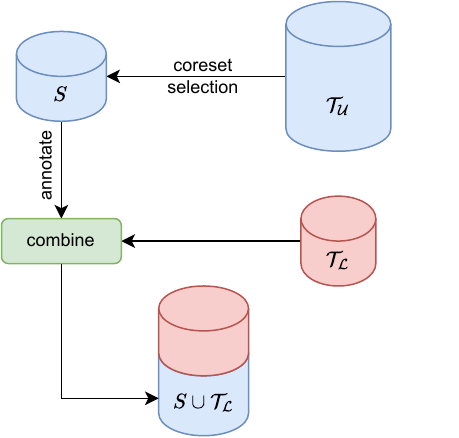}
        \caption{\label{fig:activeLearning}
        The active learning pipeline. The idea of active learning is to prune a large pool of label-free dataset $\mathcal{U}$ and to extend the already existing labeled dataset with essential, newly annotated data.
        }
    \end{center}
\end{figure}

Building on the distinction between label-aware coreset selection (where labels are available and often directly exploited) and \emph{label-free} selection (where selection is driven purely by inputs), a closely related research field is \emph{active learning}, which targets data efficiency from a partially labeled perspective \cite{sener2017active, settles2009active, loeffler2024learning}. 
Instead of selecting a coreset $\mathcal{S}$ from a fully labeled dataset, active learning assumes access to a large unlabeled pool $\mathcal{T}_\mathcal{U}$ and asks which label-free samples should be queried from an oracle for annotation (e.g., a human annotator or nearest neighbor annotator) \cite{druck2009active, huang2016active, zhang2023llmaaa}. 
Thus, the goal is to reduce the label-free pool $\mathcal{T}_\mathcal{U}$ by selecting an informative query batch $\mathcal{Q}\subset \mathcal{T}_\mathcal{U}$, as illustrated in \autoref{fig:activeLearning}.

Formally, consider a large label-free dataset $\mathcal{T}_\mathcal{U}$ and a small labeled dataset  $\mathcal{T}_\mathcal{L}$, where $\mathcal{T}_\mathcal{U} \cap \mathcal{T}_\mathcal{L} = \emptyset$. 
The goal of active learning is to iteratively augment $\mathcal{T}_\mathcal{L}$ by selecting a subset $\mathcal{S} \subset \mathcal{T}_\mathcal{U}$ such that $|\mathcal{S}| \ll |\mathcal{T}_\mathcal{U}|$ while maximizing the expected model performance on $\mathcal{S} \cup \mathcal{T}_\mathcal{L}$.
Active learning aims to solve the following optimization problem:
\begin{equation}
\mathcal{S}^* = \argmin_{\mathcal{S} \subset \mathcal{T}_\mathcal{U}, \frac{|\mathcal{S}|}{|\mathcal{T}_\mathcal{U}|} \leq 1 - \alpha} \mathbb{E}_{\mathbf{x}, y \sim P} [\mathcal{L}(\mathbf{x}, y; \theta^\mathcal{S})],
\end{equation}
where $\theta^\mathcal{S}$ is the model trained on $\mathcal{T}_\mathcal{L} \cup \mathcal{S}$,  $\alpha \in (0,1)$ is the fraction of data left label-free, and  $\mathcal{L}(\mathbf{x}, y; \theta)$  is the model loss (e.g., cross-entropy loss).

In practice, several coreset strategies, especially those focusing on “most informative” or “diverse” examples, can be adapted to active learning settings by simply applying them to label-free data and then requesting annotations \cite{ducoffe2018adversarial, kothawade2021similar}. 
Conversely, many established active learning techniques also serve as strong coreset baselines (if the samples are already labeled) \cite{wei2015submodularity, sener2017active}.

\begin{table*}[t!]
\centering
\caption{\textbf{Boundaries between related data-reduction paradigms.}
We contrast supervision assumptions, outputs, training-signal access, and primary goals.}
\label{tab:boundaries}
\renewcommand{\arraystretch}{1.15}
\setlength{\tabcolsep}{4pt}
\footnotesize
\begin{tabularx}{\textwidth}{p{0.13\textwidth} p{0.16\textwidth} p{0.14\textwidth} X X}
\toprule
\textbf{Paradigm} &
\textbf{Label assumption} &
\textbf{Output artifact} &
\textbf{Training access / signals} &
\textbf{Primary objective \& typical methods} \\
\midrule

\textbf{Coreset selection} &
Fully labeled dataset $\mathcal{T}=\{(x_i,y_i)\}_{i=1}^{N}$ &
Subset of real examples $\mathcal{S}\subset\mathcal{T}$ (optionally weighted) &
May be training-free or uses (partial) training dynamics (loss, gradients, margins) &
Match performance of training on $\mathcal{T}$ at budget $(1-\alpha)$.
\newline
Methods: Random; geometry/coverage (Herding, $k$-Center); scoring (Forgetting, GraNd/EL2N); submodular (Facility Location/GraphCut); gradient matching (CRAIG/GradMatch); bilevel (GLISTER). \\

\addlinespace[2pt]
\textbf{Active learning} &
Large label-free pool; labels queried from an oracle &
Query batch $\mathcal{Q}\subset\mathcal{T}_u$ to label; iteratively expands $\mathcal{T}_l$ &
Iterative; relies on model uncertainty/disagreement + diversity criteria &
Maximize accuracy per \emph{annotation} budget (label efficiency).
\newline
Methods: uncertainty (entropy/margin), diversity/coverage batches (coresets/submodular), disagreement/committee, boundary/adversarial-based querying. \\

\addlinespace[2pt]
\textbf{label-free coreset selection} &
No ground-truth labels available during selection &
Subset $\mathcal{S}\subset\{x_i\}_{i=1}^{N}$ (sometimes pseudo-labels/weights) &
No supervised training signal; uses representation geometry or self-supervised proxies &
Preserve downstream utility under compute/storage constraints without labels.
\newline
Methods: coverage/diversity in foundation-model feature space (CLIP/DINO + $k$-Center/submodular);
pseudo-labeling + supervised selectors (ELFS-style);
consistency/disagreement under augmentations; multimodal agreement signals. \\

\addlinespace[2pt]
\textbf{Semi-supervised coreset selection} &
Small labeled $\mathcal{T}_l$ + large label-free $\mathcal{T}_u$ &
Subset $\mathcal{S}\subset\mathcal{T}_u$ (or weighted label-free set) &
Uses labeled supervision sparsely + SSL losses (consistency/pseudo-labeling) &
Pick label-free points that best complement $\mathcal{T}_l$ for generalization.
\newline
Methods: bilevel/validation-driven (RETRIEVE), confidence filtering + diversity, hybrid submodular objectives. \\

\addlinespace[2pt]
\textbf{Dataset distillation / condensation} &
Typically labeled (or labels generated) &
Synthetic set $\tilde{\mathcal{S}}=\{(\tilde{x}_j,\tilde{y}_j)\}_{j=1}^{M}$, $M\ll N$ &
Heavy optimization; matches gradients/trajectories/distributions of training on $\mathcal{T}$ &
Replace real data with a compact \emph{synthetic} training set.
\newline
Methods: gradient/trajectory matching (MTT), distribution matching, bilevel objectives, generative priors; coreset selection can be pre-/post-processing. \\

\bottomrule
\end{tabularx}
\end{table*}

\subsection{Coverage}
An often overlooked property, especially for earlier coreset selection methods, is coverage \cite{zheng2022coverage, indyk2014composable}.
Intuitively, we describe a coreset with good coverage as a set representing the whole range of the dataset distribution, not only the easiest/hardest samples.  A well-selected coreset balances the original dataset's geometric coverage while preserving its structural diversity. Every point in $\mathcal{T}$ is geometrically or in a feature space close to at least one point in a well-selected coreset $\mathcal{S}$. More formally, we can quantify (percentage-wise) coverage $c$ in the following way:

\begin{equation}
    c = \int_{X} \mathbf{1}_\mathrm{\cup_{\mathbf{x} \in \mathcal{S}}B_d(\mathbf{x}, r)}(\mathbf{x}) d\mathbb{P}(\mathbf{x}),
\end{equation}
where $B_d(\mathbf{x}, r) = \{\mathbf{x}' \in X: d(x, x') \leq r \}$ is a $r$-radius ball whose center is $\mathbf{x}$,  and $d(\cdot, \cdot)$ is a distance metric.

\section{Training-Free Methods}
\label{sec:training-free}
In this section, we introduce training-free methods, namely, random selection and geometry-based selection.
\subsection{Random Selection}
A straightforward training-free baseline is to select each sample from $\mathcal{T}$ with uniform probability:
\begin{equation}
    \mathbb{P} \left[(\mathbf{x}, y) \in S\right] = \frac{|\mathcal{S}|}{|\mathcal{T}|} = 1 - \alpha. 
\end{equation}

Despite its simplicity, random sampling remains an baseline reference in coreset research for two reasons. First, its computational minimalism requires zero overhead beyond the selection step, i.e., no model training, no extra distance computations, etc.
Second, in over-parameterized settings or for large, diverse datasets, even a random coreset can perform competitively \cite{gupta2023data, guo2022deepcore}. 
Redundant examples in $\mathcal{T}$ indicate that a randomly chosen $\mathcal{S}$ often covers a substantial fraction of the data distribution \cite{li2023exploiting, birodkar2019semantic}.

Another benefit of random sampling is that it is extremely fast, trivially parallelizable, and unbiased with respect to class or difficulty.
However, the latter is also a disadvantage, as it ignores data distribution as well as sample difficulty and tends to become suboptimal when the pruning ratio $\alpha$ increases (i.e., above 90\%).

\subsection{Geometry Based}
A fundamental and well-known assumption in machine learning is that samples closer in feature space tend to share similar properties \cite{caron2018deep,hastie2017elements}.
Geometry-based methods exploit this intuition by identifying redundant or noninformative samples while selecting a subset that effectively represents the original data distribution \cite{bachem2018scalable}. 

\subsubsection{Herding}
\begin{figure}[t!]
    \begin{center}
        \includegraphics[width=.48\textwidth]{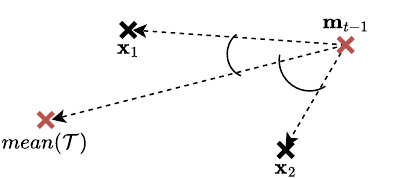}
        \caption{\label{fig:herding}
        Illustration of the iterative selection of Herding. As the angle between the direction to the mean of the real dataset $\mathcal{T}$ and to $\mathbf{x}_1$ is smaller than to $\mathbf{x}_2$, as well as the distance to $\mathbf{m}_{t-1}$ is higher to $\mathbf{x}_1$, $\mathbf{x}_1$ will be included for the iteration step $t$, leading to a new coreset mean $\mathbf{m}_t$.
        }
    \end{center}
\end{figure}

Herding sequentially selects samples that best preserve the statistical properties of the full dataset \cite{welling2009herding, chen2012super}. 
Unlike random sampling, Herding greedily selects points to minimize the discrepancy between the coreset mean and the dataset mean in feature space, called \textit{moment constraint} \cite{kalischek2021light, liu2021sampling}.
Specifically, given a feature mapping $\phi: \mathcal{X} \to \mathbb{R}^d$, the moment constraint requires that the empirical mean of the selected coreset approximates that of the full dataset:

\begin{equation}
    \frac{1}{|\mathcal{T}|} \sum_{\mathbf{x} \in \mathcal{T}} \phi(\mathbf{x}) \approx \frac{1}{|\mathcal{S}|} \sum_{\mathbf{x} \in \mathcal{S}} \phi(\mathbf{x}).
\end{equation}
\noindent
As shown in \autoref{fig:herding}, Herding iteratively selects the next sample $\mathbf{x}_t$ by minimizing the deviation between these two means.

\begin{equation}
    \mathbf{x}_{t} = \arg\max_{\mathbf{x} \in \mathcal{T}} \langle \phi(\mathbf{x}), \mathbf{m}_{t-1} \rangle,
\end{equation}
\noindent
where $\mathbf{m}_t = \mathbf{m}_{t-1} + \phi(\mathbf{x}_t) - \frac{1}{|\mathcal{S}|} \sum_{\mathbf{x} \in \mathcal{S}} \phi(\mathbf{x})$ represents the running mean update and $\langle \mathbf{a}, \mathbf{b} \rangle$ is the inner product of two vectors, which measures their alignment.

One benefit of Herding is that it is easy to implement if a suitable embedding $\phi(\cdot)$ is available.
While it preserves the global statistics (the mean), its focus may overlook more subtle distribution patterns (e.g., multi-modal clusters). 
Moreover, it applies iterative updates, which can be costly for very large $\mathcal{T}$.

\subsubsection{k-Center Greedy}
Another popular geometry-based method is k-Center Greedy, which aims to minimize the maximum distance between \emph{any} point in $\mathcal{T}\setminus\mathcal{S}$ and the “closest” point in $\mathcal{S}$ \cite{sener2017active}. 
Formally, it solves the \textit{minimax facility location} problem \cite{farahani2009facility}:

\begin{equation}
\min_{\mathcal{S} \subset \mathcal{T}, \frac{|\mathcal{S}|}{|\mathcal{T}|}=(1-\alpha)} \quad \max_{\mathbf{x}_i \in \mathcal{T} \setminus \mathcal{S}} \quad \min_{\mathbf{x}_j \in \mathcal{S}} d(\mathbf{x}_i, \mathbf{x}_j),
\end{equation}
\noindent
where  $d(\mathbf{x}_i, \mathbf{x}_j)$ is a predefined distance metric.
The goal is to select $k = (1-\alpha) \cdot {|\mathcal{T}|}$ points that serve as centers, ensuring that every remaining point is close to at least one of these centers. 
Since solving this problem exactly is NP-hard, an efficient greedy approximation is typically used \cite{sener2017active}.

The $k$-Center Greedy method can be interpreted as finding a set of representative samples that provide maximal coverage of the dataset under a geometric distance measure. 
It is commonly used in active learning and coreset selection since it ensures that no part of the dataset is too far from the selected subset \cite{agarwal2020contextual, sinha2020small}.
Unlike Herding, which aims to match the dataset embedding mean by enforcing moment constraints, $k$-Center Greedy focuses purely on geometric coverage, targeting a more diverse and well-distributed coreset.
However, its strict focus on coverage might not align perfectly with model-specific decision boundaries.
Moreover, its computation time is considerably high compared to other methods.

\subsection{Embedding Space}
\begin{figure}[t!]
    \begin{center}
        \includegraphics[width=.48\textwidth]{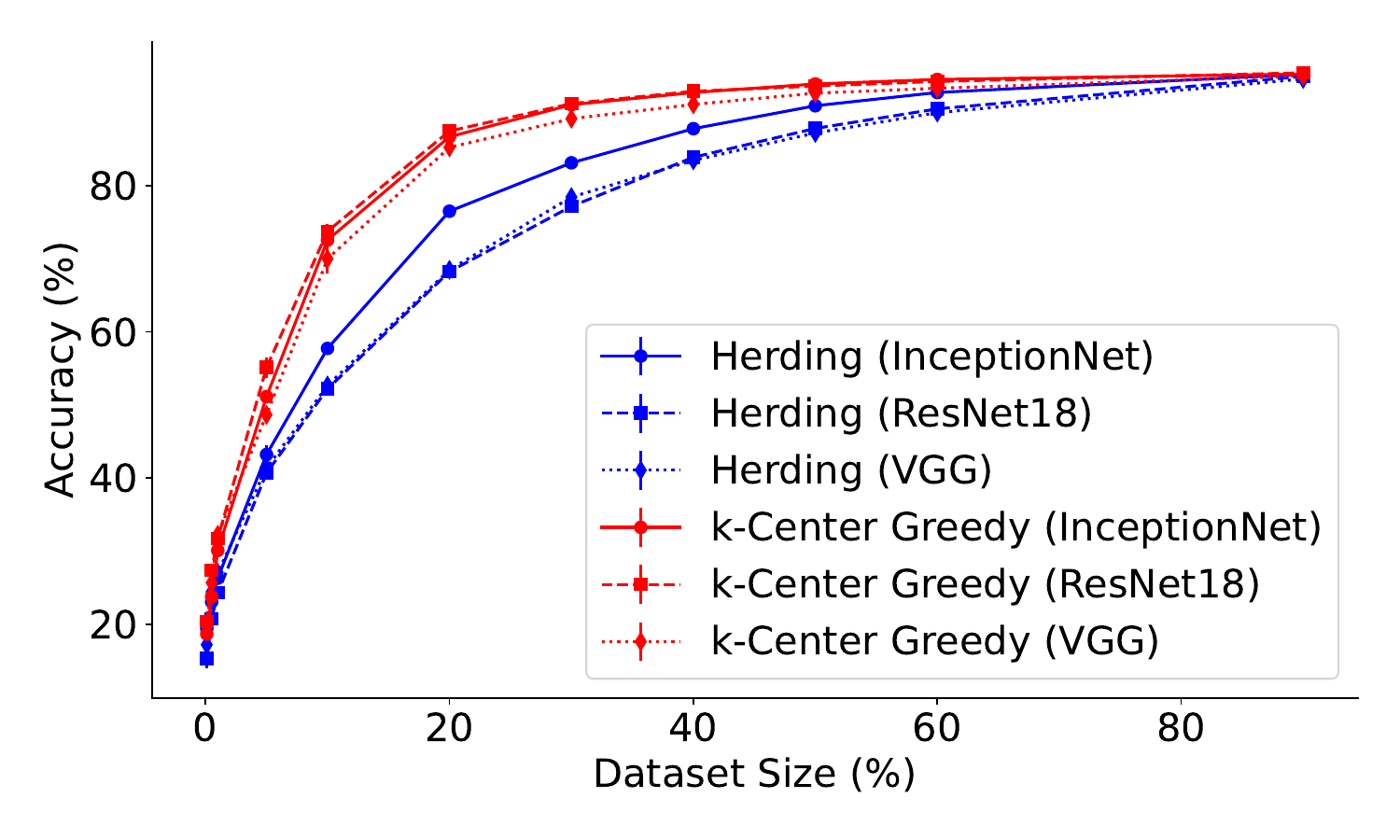}
        \caption{\label{fig:herding_kcenter}
        Coreset selection performance of Herding and k-Center Greedy on CIFAR-10 for different feature embeddings (ResNet-18 \cite{he2016deep}, InceptionNet \cite{szegedy2015going}, and VGG \cite{simonyan2014very}).
        }
    \end{center}
\end{figure}

As shown in \autoref{fig:herding_kcenter}, geometry-based methods like Herding and k-Center Greedy heavily rely on the embedding quality of the underlying pre-trained feature extractor.
A recent shift is to replace task-trained backbones (e.g., supervised ImageNet encoders) with \emph{foundation-model} representations that are substantially more transferable and semantically aligned across domains \cite{yang2024clip, gomez2025reducing, wang2025efficient}.
In vision, strong options include large-scale self-supervised ViTs such as DINO/DINOv2~\cite{caron2021emerging, oquab2023dinov2} and vision-language pretraining such as CLIP-style models~\cite{radford2021learning}.
In practice, this turns many classical training-free selectors into \emph{plug-and-play} methods: given an encoder $\phi$, we embed each image
$
\mathbf{z}_i = \phi(\mathbf{x}_i)
$
and run the usual selection objective (e.g., $k$-Center or Herding) directly in the representation space.

With stronger embeddings, distance/similarity becomes a better proxy for semantic redundancy and coverage.
This perspective also unifies \emph{training-free} selection with recent label-free works: they often differ mainly in which encoder is used (self-supervised vs.\ vision-language) and whether additional proxies (e.g., pseudo-labels or difficulty estimates) are layered on top.

\section{Training-Oriented Methods}
\label{sec:training-oriented}
One way to circumvent the embedding space problem is to train the model on the target dataset that needs to be pruned.
However, when training on the target dataset is required, one can use more sophisticated methods that also take into account the training dynamics, which fall under the umbrella term of training-oriented methods.

By tailoring a subset $\mathcal{S}$ to the actual training dynamics, these methods can often achieve higher accuracy than their training-free counterparts, albeit at the expense of additional computation.
In this section, we present five major families of training-oriented coreset strategies. 
These include (i) scoring-based, (ii) decision boundary-based, (iii) submodularity-based, (iv) gradient matching-based, and (v) bilevel optimization-based methods. 
Throughout, we highlight how each method leverages partial or full training signals. 

\begin{figure}[t!]
    \begin{center}
        \includegraphics[width=.75\columnwidth]{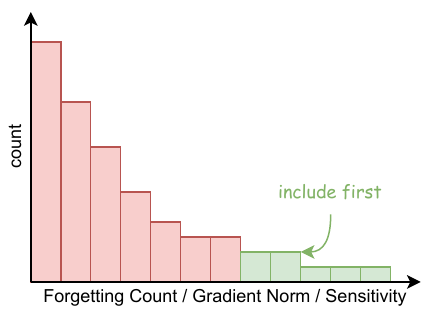}
        \caption{\label{fig:loss_value_based}
        Illustration of scoring-based methods. Usually, they employ a metric derived from multiple training iterations that quantifies the impact of a sample on the training quality (e.g., with forgetting counts, gradient norms, or sensitivity scores). The coreset is then selected by including the highest scores first. Note that the typical power law distribution of scores is only found for sufficiently trained and parameterized neural networks that are used for deriving the scores. Under-parameterized neural networks would lead to a more uniform distribution. Further details can be found in our theoretical remarks in \autoref{sec:theoretical-remarks}.
        }
    \end{center}
\end{figure}

In the following, we use \emph{importance} as an umbrella term for a sample-specific \emph{utility proxy}, a measurable signal that approximates the benefit of including $(\mathbf{x},y)$ under a fixed selection budget (typically in terms of generalization).
Different paradigms instantiate this proxy differently: scoring-based methods estimate \emph{training influence} via per-sample dynamics (loss/forgetting/gradient magnitude), submodular methods define \emph{set-level} utilities that encourage coverage and non-redundancy, and gradient-matching/bilevel methods align selection with \emph{optimization} or \emph{validation} objectives.
Thus, “importance” should be read as \emph{proxy-dependent}, not as a single universal quantity.

\subsection{Scoring-Based}

A natural way to incorporate training knowledge is to assign each sample a \emph{score} that reflects its importance to the learning process. 
We then construct the coreset by selecting data with the highest scores. 
Scores can come from forgetting events, gradient norms, or importance sampling criteria, as illustrated in \autoref{fig:loss_value_based}.

\subsubsection{Forgetting}
Capturing forgetting events is beneficial to analyze the stability of sample classifications throughout the training \cite{toneva2018empirical}. 
A forgetting event occurs when a sample, correctly classified at some time step $t$, is later misclassified at $t{\prime} > t$. 

Formally, let $y_i^t = \argmax p(\mathbf{x}_i; \theta)$  be the predicted label of sample \( (\mathbf{x}_i, y_i) \) at epoch $t$.  
A forgetting event is counted whenever:

\begin{equation}
    \mathbf{1}_{(y_i^t = y_i)} = 1 > \mathbf{1}_{(y_i^{t+1} = y_i)} = 0,
\end{equation}
where $\mathbf{1}$ is the indicator function.

This metric can categorize samples as forgettable (i.e., experiencing one or more forgetting events) or unforgettable (i.e., never forgotten once learned).
As a result, we can rank samples based on the number of forgetting events.
Empirical studies suggest that unforgettable examples are often well-separated from the decision boundary \cite{swayamdipta2020dataset, toneva2018empirical, hooker2019compressed}. 
This separation typically indicates redundancy.
Forgettable examples, on the other hand, contribute more to refining the decision boundary and generalization.
Thus, coreset selection methods based on forgetting take samples with the highest forgetting counts first.

One benefit of utilizing forgetting is that it is easy to implement: just count forgetting events over training epochs.
However, it can be highly nondeterministic due to its reliance on factors like optimization schedules (e.g., learning rate decay) and random seeds \cite{swayamdipta2020dataset}.

\subsubsection{GraNd}
Another method of scoring samples is to use the Gradient Norm (GraNd), which measures the contribution of each sample to the loss function \cite{paul2021deep}.
Similar to forgetting, we exploit the insight of a training trajectory. 
Given an epoch $t$, the GraNd score is defined as 
\begin{equation}
\chi_t(\mathbf{x}, y) = \mathbb{E}_{\theta_t} \|\nabla{\theta_t} \mathcal{L}(\mathbf{x}, y; \theta_t)\|_2, 
\end{equation}
\noindent
where larger values indicate higher importance, as they indicate a greater force on parameter updates.
Thus, we select samples with larger values first.

To reduce the computational cost, an alternative approximation, the Expected L2 Norm (EL2N), replaces the gradient computation with the squared Euclidean norm of the error vector in the output space:
\begin{equation}
    \chi^*_t(\mathbf{x}, y) = \mathbb{E}_{\theta_t} \| p(\mathbf{x}; \theta_t) - y \|_2^2,
\end{equation}
\noindent
where  $p(\mathbf{x}; \theta_t)$  is the model prediction (i.e., a probability distribution). 
This approximation avoids backpropagation through the entire network, which was trained beforehand, while still capturing the difficulty of learning a sample.

The benefit of GraNd/EL2N is that it directly relates to the extent to which a sample drives parameter updates. 
Moreover, they are straightforward ranking metrics once partial training is done.
However, similar to GraNd, extremely high-sensitivity samples (e.g., outliers) can dominate the sampling distribution \cite{alain2015variance,citovsky2021batch}. 
This skew risks overfitting and can reduce coverage of the broader data distribution.
Among all the methods reviewed, we find that GraNd has the highest runtime.

\subsubsection{Sensitivity}
Another importance measure for coreset selection is the sensitivity of a sample, which quantifies its worst-case contribution to the total loss function across model parameters $\theta_t$ for multiple epochs $t$:

\begin{equation}
    s(\mathbf{x}, y) = \max_{\theta_t} \frac{\mathcal{L}(\mathbf{x}, y; \theta_t)}{\sum_{(\mathbf{x}{\prime}, y{\prime}) \in \mathcal{T}} \mathcal{L}(\mathbf{x}{\prime}, y{\prime}; \theta_t)}.
\end{equation}
\noindent
The sampling probability for each sample is then
\begin{equation}
    \mathbb{P} \left[(\mathbf{x}, y) \in S\right] = \frac{s(\mathbf{x}, y)}{\sum_{(\mathbf{x}{\prime}, y{\prime}) \in \mathcal{T}} s(\mathbf{x}{\prime}, y{\prime})},
\end{equation}
\noindent
i.e., samples with higher sensitivity are more likely to be included in $\mathcal{S}$ as they are critical to preserving structure, while redundant or less informative samples have lower selection probabilities \cite{bachem2015coresets, munteanu2018coresets}.

Importance sampling with sensitivity comes from well-established principles in stochastic optimization \cite{zhao2015stochastic}. 
It has a clear interpretation in terms of weighting samples by their expected contribution. 
However, similar to GraNd, if certain samples have extremely high sensitivity (e.g., outliers), sampling can become heavily skewed toward a small subset, risking overfitting or ignoring coverage of the broader data distribution \cite{alain2015variance,citovsky2021batch}. 
Moreover, computing exact sensitivities can be computationally expensive, especially for complex models or high-dimensional data.
Also, it can require careful tuning of temperature or smoothing parameters to avoid collapsing on a small subset of high-loss samples \cite{NEURIPS2018, zhao2015stochastic}.

\subsection{Decision Boundary Based}
Whereas scoring-based approaches track \emph{loss signals} over time, decision boundary-based methods explicitly estimate proximity to the classification boundary \cite{yang2024mind, cho2025lightweight}. 
These methods assume that samples near the boundary are more informative, i.e., have a higher importance.

\begin{figure}[t!]
    \begin{center}
        \includegraphics[width=.75\columnwidth]{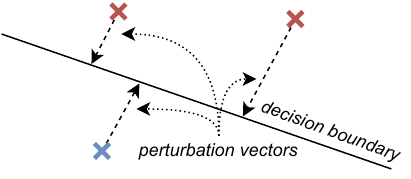}
        \caption{\label{fig:deepfool}
        Illustration of DeepFool. It calculates for each data sample a perturbation $\mathbf{r}$ and selects the samples with the shortest perturbations first (minimal norm). Thus, it prioritizes data near the decision boundary.
        }
    \end{center}
\end{figure}

\begin{figure}[t!]
    \begin{center}
        \includegraphics[width=.8\columnwidth]{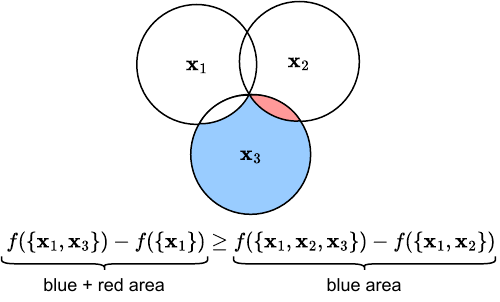}
        \caption{\label{fig:submodul}
        Example of a submodular function where $A= \{\mathbf{x}_1\} \subset B = \{\mathbf{x}_1, \mathbf{x}_2\}$: $B_d(\mathbf{x}, r) = |\{\mathbf{x}' \in X: d(x, x') \leq r \}|$, where $B_d(\mathbf{x}, r)$ is a $r$-radius ball whose center is $\mathbf{x}$. Naturally, a smaller set of centers has a higher chance of encapsulating more distinguished points.
        }
    \end{center}
\end{figure}

\subsubsection{DeepFool}
DeepFool approximates a sample’s distance to the decision boundary using adversarial perturbations \cite{ducoffe2018adversarial, moosavi2016deepfool}. 
In this sense, it is a geometry-based selector.
The key idea is that samples near the decision boundary carry more information for classification than those farther away, as illustrated in \autoref{fig:deepfool}.

Given a trained classifier $p(\mathbf{x}; \theta)$, DeepFool iteratively finds the smallest perturbation $\mathbf{r}$ needed to move a sample $\mathbf{x}$ across the decision boundary:
\begin{align}
    \mathbf{x}{\prime} = \mathbf{x} + \mathbf{r},  \text{ where } \mathbf{r} = \argmin_{\mathbf{r}{\prime}} \|\mathbf{r}{\prime}\|_2 \nonumber\\ 
    \text{ s.t. } \argmax p(\mathbf{x} + \mathbf{r}; \theta) \neq \argmax p(\mathbf{x}; \theta).
\end{align}
\noindent
For coreset selection, DeepFool prioritizes samples with the smallest perturbation magnitude $\|\mathbf{r}\|_2$, since they are closest to the decision boundary. 
%
In contrast to training-free geometry-based methods that do not account for the actual learned boundary, DeepFool often yields improved performance.
However, it requires repeated adversarial steps, which can be expensive and sensitive to the adversarial hyperparameters (e.g., step size).

\subsubsection{Contrastive Active Learning}
Contrastive Active Learning (CAL) balances uncertainty sampling and coverage sampling by selecting contrastive examples \cite{margatina2021active, chen2020simple, deng2023counterfactual}. 
These are points that are close in feature space but yield highly divergent model predictions.
This approach ensures that selected examples contribute maximally to refining the decision boundary.

Specifically, CAL defines a contrastive pair $(\mathbf{x}_i, \mathbf{x}_j)$ based on two criteria: Feature similarity and predictive discrepancy.
Regarding the first, CAL is optimized to map the data points to be close in the feature space:
\begin{equation}
    d(\Phi(\mathbf{x}_i), \Phi(\mathbf{x}_j)) < \epsilon
\end{equation}
\noindent
where $\Phi(\cdot)$ is an encoder mapping input samples to a shared feature space, $d(\cdot, \cdot)$ is a distance metric, and $\epsilon > 0$ is a small threshold.
With respect to predictive discrepancy, the model should assign significantly different predictive probabilities to the two examples:
\begin{equation}
    D_{\text{KL}}( p(y | \mathbf{x}_i) \, || \, p(y | \mathbf{x}_j) ) \to \infty,
\end{equation}
where $D_{\text{KL}}$ denotes the Kullback-Leibler (KL) divergence.
Intuitively, contrastive examples are samples that lie near the decision boundary, where the model is uncertain. 
CAL prioritizes these samples for the coreset.

One benefit of CAL is that it balances data diversity (i.e., closeness in feature space) with uncertainty (i.e., mismatched predictions).
Nevertheless, it also requires computing pairwise distances and KL divergences, which can be costly at scale.
Similar to geometry-based, training-free approaches, the performance of CAL also depends heavily on the quality of the learned embedding.

\subsection{Submodularity Based}
Submodularity is a fundamental property of set functions that captures the principle of diminishing returns, which makes it especially interesting for coreset selection \cite{iyer2013submodular, kothawade2021submodular, karanam2022orient, wei2015submodularity, doucoreset}. 

A function  $f: 2^V \to \mathbb{R}$  defined over a ground set $V$ is submodular if, for any subsets  $A \subseteq B \subseteq V$  and any element  $j \in V \setminus B$ , the following holds:
\begin{equation}
    f(A \cup \{j\}) - f(A) \geq f(B \cup \{j\}) - f(B).
\end{equation}
\noindent
This property, known as the \textit{diminishing returns condition}, means that adding an element to a smaller set provides a greater marginal gain than adding it to a larger set. 
\autoref{fig:submodul} shows an example.
Intuitively, submodular functions model concepts like diversity, coverage, and representativeness.

In general, our goal in coreset selection is to maximize a submodular function:
\begin{equation}
    \mathcal{S}^* = \argmax_{\mathcal{S} \subseteq V, \frac{|\mathcal{S}|}{|\mathcal{T}_u|} \leq 1 - \alpha} f(\mathcal{S}).
\end{equation}
Common submodular functions in coreset selection are Facility Location (maximizing coverage) with
\begin{equation}
    f(\mathcal{S}) = \sum_{\mathbf{x} \in \mathcal{T}} \max_{\mathbf{x}{\prime} \in \mathcal{S}} \text{sim}(\mathbf{x}, \mathbf{x}{\prime}),
\end{equation}
where $\text{sim}(\mathbf{x}, \mathbf{x}{\prime})$ is a similarity metric (e.g., cosine similarity) and Graph Cut (minimizing redundancy) with
\begin{equation}
    f(\mathcal{S}) = \sum_{\mathbf{x} \in \mathcal{S}} \sum_{\mathbf{x}{\prime} \in \mathcal{S}} w(\mathbf{x}, \mathbf{x}{\prime}),
\end{equation}
\noindent
where $w(\mathbf{x}, \mathbf{x}{\prime})$ represents a similarity weight.

However, this problem is NP-hard, meaning that finding the exact optimal subset $\mathcal{S}^*$  is computationally infeasible for large datasets.
Luckily, submodular functions exhibit a key optimization advantage: they can be efficiently maximized using a greedy algorithm, which provides a  $(1 - 1/e) \approx 63\%$ approximation guarantee for cardinality-constrained maximization \cite{nemhauser1978analysis}. 
This guarantee means that if $f(\mathcal{S}^*)$ is the best possible value, then the greedy-selected subset  $\mathcal{S}_\text{greedy}$ satisfies:
\begin{equation}
    f(\mathcal{S}_\text{greedy}) \geq (1 - 1/e) \cdot f(\mathcal{S}^*).
\end{equation}
This tells us a few things: 1) Greedy methods give strong approximations without needing brute-force optimization.
2) The greedy algorithm will never be worse than about 63\% of the best possible solution.
3) Lazy greedy techniques can be employed to speed up computation significantly \cite{lim2014lazy}.

A naturally arising question is: Why is it not 100\% optimal? 
The reason we cannot achieve exactly  $f(\mathcal{S}^*)$ with a greedy method is that each step picks the locally best option without accounting for global interactions in the dataset. 
This is known as \textit{greedy suboptimality}.
However, for most practical submodular maximization problems (including coreset selection), the greedy algorithm with a $(1 - 1/e)$ guarantee is a well-established strong baseline.

\subsubsection{FASS}
Filtered Active Submodular Selection (FASS) is a method designed to improve active learning by combining uncertainty sampling with submodular optimization \cite{wei2015submodularity}. 
The key idea is to filter uncertain examples first and then select the most representative subset using submodular optimization.

Consider a dataset $\mathcal{T} = \mathcal{T}_l \cup \mathcal{T}_u$  consisting of labeled ($\mathcal{T}_l$) and label-free ($\mathcal{T}_u$) samples.
FASS first selects a candidate pool $\mathcal{C}$ of uncertain examples from $\mathcal{T}_u$. 
A typical uncertainty measure is entropy:
\begin{equation}
    H(\mathbf{x}) = -\sum_{c} p(y_c | \mathbf{x}; \theta) \log p(y_c | \mathbf{x}; \theta),
\end{equation}
\noindent
where $ p(y_c | \mathbf{x}; \theta)  $ is the model prediction probability of class c.
The candidate pool $\mathcal{C}$ is then created from the top $m$ most uncertain samples:
\begin{equation}
    \mathcal{C} = \argmax_{\mathcal{C} \subseteq \mathcal{T}_u, |\mathcal{C}|=m} \sum_{\mathbf{x} \in \mathcal{C}} H(\mathbf{x}).
\end{equation}
Given the candidate set $\mathcal{C}$, we select a batch $\mathcal{S}$ of size $(1-\alpha) \cdot |\mathcal{T}|$ by maximizing a submodular function $f$:
\begin{equation}
    \mathcal{S}^* = \argmax_{\mathcal{S} \subseteq \mathcal{C}, \frac{|\mathcal{S}|}{|\mathcal{T}|} = (1-\alpha)} f(\mathcal{S}).
\end{equation}

The key benefit of FASS is that it reduces the computational cost by restricting submodular selection to uncertain points, potentially merging the best of both worlds (uncertainty and coverage).
Yet, the computational costs are still considerably higher than those of other coreset selection methods.

\subsubsection{PRISM}
In a similar vein, PRISM (Parameterized Submodular Information Measures) builds on submodular information measures, namely CG, MI, and CMI \cite{kothawade2022prism}. 
In the following, we will explain each measure in detail.
The submodular Conditional Gain (CG) measures the gain in function value when adding a set $\mathcal{S}$ to a validation set $\mathcal{V} = \mathcal{T} \setminus \mathcal{S}$:
\begin{equation}
    f(\mathcal{S} | \mathcal{V}) = f(\mathcal{S} \cup \mathcal{V}) - f(\mathcal{V}).
\end{equation}
CG is used to quantify how different a selected set $\mathcal{S}$ is from $\mathcal{V}$.
On the other hand, the Submodular Mutual Information (MI) captures the similarity between sets $\mathcal{S}$ and $\mathcal{T}$:
\begin{equation}
    I_f(\mathcal{S}; \mathcal{T}) = f(\mathcal{S}) + f(\mathcal{T}) - f(\mathcal{S} \cup \mathcal{T})
\end{equation}
MI encourages selecting elements similar to $\mathcal{T}$.
By combining both, we obtain Submodular Conditional Mutual Information (CMI), which jointly models both the similarity to $\mathcal{T}$ and dissimilarity from $\mathcal{V}$:
\begin{equation}
    I_f(\mathcal{S}; \mathcal{T} | \mathcal{V}) = f(\mathcal{S} | \mathcal{V}) + f(\mathcal{T} \cup \mathcal{V}) - f(\mathcal{S} \cup \mathcal{T} \cup \mathcal{V})
\end{equation}
Thus, the overall optimization goal for a subset is 
\begin{equation}
    \mathcal{S}^* = \argmax_{\mathcal{S} \subseteq \mathcal{T}, \frac{|\mathcal{S}|}{|\mathcal{T}_u|} \leq 1 - \alpha} I_f(\mathcal{S}; \mathcal{T} | \mathcal{V}).
\end{equation}

PRISM shares the advantages and disadvantages of FASS. 
Still, the CMI-based framework unifies the notion of coverage (similar to facility location) with a penalty for being too similar to existing sets.

\subsubsection{SIMILAR}
SIMILAR (Submodular Information Measures Based Active Learning In Realistic Scenarios) extends submodular coreset selection by leveraging Submodular Mutual Information (SMI) and Submodular Conditional Mutual Information (SCMI) to select diverse and informative subsets \cite{kothawade2021similar}. 
The framework unifies uncertainty sampling, diversity-based selection, and coverage-based selection within a single mathematical formulation.

SMI quantifies the mutual dependence between a selected coreset $\mathcal{S}$ and a reference set $\mathcal{Q}$ (which could be the full dataset, rare-class samples, or out-of-distribution data) using:
\begin{equation}
I_f(\mathcal{S}; \mathcal{Q}) = f(\mathcal{S}) + f(\mathcal{Q}) - f(\mathcal{S} \cup \mathcal{Q}),
\end{equation}
where $f(\cdot)$ is a submodular function modeling representativeness or diversity.

To further refine the selection process, SIMILAR introduces SCMI, which penalizes redundancy by conditioning on an already selected set $\mathcal{V}$:
\begin{equation}
I_f(\mathcal{S}; \mathcal{Q} | \mathcal{V}) = f(\mathcal{S} | \mathcal{V}) + f(\mathcal{Q} | \mathcal{V}) - f(\mathcal{S} \cup \mathcal{Q} | \mathcal{V}).
\end{equation}

This enables a more context-aware selection strategy, ensuring that selected samples are informative relative to both the dataset $\mathcal{Q}$ and the existing knowledge in $\mathcal{V}$.

The coreset selection follows:
\begin{equation}
\mathcal{S}^* = \argmax_{\mathcal{S} \subseteq \mathcal{T}, \frac{|\mathcal{S}|}{|\mathcal{T}|} \leq 1 - \alpha} I_f(\mathcal{S}; \mathcal{Q} | \mathcal{V}),
\end{equation}
which allows for adaptive batch selection, ensuring diversity and non-redundancy.

SIMILAR provides a generalized coreset selection framework applicable to rare-class selection, out-of-distribution data filtering, and uncertainty-aware sampling. 
By leveraging lazy greedy optimization, it maintains a $(1 - 1/e)$ approximation guarantee while being more scalable than traditional submodular methods. 
However, its reliance on computing large similarity matrices increases its complexity, making it less efficient for very large datasets unless approximate submodular maximization techniques are used.

\subsection{Gradient Matching Based}
In contrast to previous approaches, gradient matching focuses on reconstructing the total gradient of the full dataset using only a subset of samples. 
This ties coreset selection directly to the training optimization path.

\subsubsection{CRAIG}
CRAIG (Coresets for Accelerating Incremental Gradient Descent) is a coreset selection method designed to select a subset of data points whose weighted gradients closely approximate the gradient of the full dataset \cite{mirzasoleiman2020coresets}.
The optimization target is to find the smallest subset $\mathcal{S}$ and associated step sizes $\gamma_j > 0$ such that:
\begin{equation}
    \max_{\theta \in W} \left\| \sum_{i \in \mathcal{T}} \nabla_{\theta} \mathcal{L}(\mathbf{x}_i, y_i; \theta) - \sum_{j \in \mathcal{S}} \gamma_j \nabla_{\theta} \mathcal{L}(\mathbf{x}_j, y_j; \theta) \right\| \leq \epsilon.
\end{equation}

Directly solving this optimization is computationally infeasible.
However, the problem can be transformed into a submodular facility location function, allowing a greedy algorithm to efficiently approximate the optimal subset.
We can incrementally build $\mathcal{S}$ by iteratively adding the sample that maximizes the marginal gain in gradient approximation. 
Each sample $\mathbf{x}_j \in \mathcal{S}$ is assigned a weight $\gamma_j$, which is determined based on the number of samples in $\mathcal{T}$ that are closest to it in the gradient space. 

The selected subset matches the full-dataset gradient well, accelerating training while maintaining performance.
Yet, it requires partial training (or iteratively updated gradients) to identify $\gamma_j$ weights, which is an additional overhead.

\subsubsection{GradMatch}
Similar in spirit is GradMatch \cite{killamsetty2021grad}. 
However, instead of ranking samples in isolation, GradMatch iteratively constructs a coreset by minimizing the gradient reconstruction error, ensuring that the subset provides a globally representative gradient direction.
Mathematically, the objective is to minimize the gradient matching error:
\begin{align}
&\mathcal{E}(\mathcal{S}) = \lambda \left\|\gamma\right\|^2 + \\
&\left\| \sum_{\mathbf{x}_i \in \mathcal{T}} \nabla_{\theta} \mathcal{L}(\mathbf{x}_i, y_i; \theta) - \sum_{\mathbf{x}_j \in \mathcal{S}} \gamma_j \nabla_{\theta} \mathcal{L}(\mathbf{x}_j, y_j; \theta) \right\|^2, \nonumber
\end{align}

GradMatch employs Orthogonal Matching Pursuit (OMP) \cite{elenberg2018restricted} to iteratively select the subset $\mathcal{S}$ by maximizing the gradient similarity. 
At each step  $t$, the algorithm selects the sample $\mathbf{x}_t$ that best reduces the gradient matching error:

\begin{equation}
\mathbf{x}_{t} = \argmax_{\mathbf{x}_i \in \mathcal{T} \setminus \mathcal{S}} \left\| \langle \nabla_{\theta} \mathcal{L}(\mathbf{x}_i, y_i; \theta), \mathbf{r}_{t-1} \rangle \right\|,
\end{equation}
where $\mathbf{r}_t$ is the residual gradient:
\begin{align}
\mathbf{r}_0 &= \sum_{\mathbf{x}_i \in \mathcal{T}} \nabla_{\theta} \mathcal{L}(\mathbf{x}_i, y_i; \theta), \text{ and}\nonumber \\
\mathbf{r}_{t} &= \mathbf{r}_{t-1} - \gamma_t \nabla_{\theta} \mathcal{L}(\mathbf{x}_t, y_t; \theta).
\end{align}

GradMatch achieves direct alignment with optimizer updates and often yields impressive compression without accuracy loss.
Nevertheless, it comes with potentially high costs for large datasets, as each step involves searching $\mathcal{T}\setminus \mathcal{S}$.
In practice, GradMatch shows sensitivity to hyperparameters (e.g., number of OMP iterations) \cite{killamsetty2021grad, elenberg2018restricted}.

In summary, gradient-matching methods are \emph{training-surrogate} selectors: they choose $\mathcal{S}$ so that one (or a few) gradient steps on $\mathcal{S}$ behaves like a gradient step on $\mathcal{T}$, i.e., they approximate the \emph{training update direction}.
In practice, this means they mainly require access to per-sample gradients under the current model and do \emph{not} need a held-out validation signal.
As a result, they often excel at \emph{speeding up optimization} (fewer examples per step) but are not explicitly optimized for downstream generalization.
Any generalization gains are indirect and depend on how well training-gradient reconstruction correlates with validation performance.

\subsection{Bilevel Optimization Based}
Bilevel optimization frameworks treat coreset selection as \emph{model selection for generalization}: the subset $\mathcal{S}$ is chosen to directly improve performance on a separate \emph{outer} objective (typically validation loss or labeled-set loss), while the model parameters are obtained by standard training on $\mathcal{S}$ in the \emph{inner} problem.
This makes bilevel methods practically different from gradient matching in two ways.
First, the selection signal is \emph{not} “does $\mathcal{S}$ reproduce the training gradient?”, but “does training on $\mathcal{S}$ reduce a held-out/generalization loss?”.
Second, they typically require either (i) an explicit validation split, or (ii) a labeled subset whose loss acts as a proxy for generalization (as in semi-supervised setups), and they use hypergradient-style approximations (e.g., one-step or truncated inner training) to score candidates.
Consequently, bilevel methods tend to be more robust to redundancy and can target distribution shift or noise via the outer objective, but they are more sensitive to how the validation/proxy set is constructed and incur higher overhead due to repeated proxy training \cite{yang2023towards, zhou2022probabilistic, xia2023refined}.

\subsubsection{RETRIEVE}
\label{retrieve}
The RETRIEVE method \cite{killamsetty2021retrieve} extends gradient-based coreset selection to the Semi-Supervised Learning (SSL) setting \cite{schmarje2021survey}, where both labeled and label-free data must be leveraged effectively. 
Unlike GradMatch, which formulates coreset selection as a gradient matching problem in a fully supervised setting, RETRIEVE introduces a bilevel optimization framework that selects the most informative label-free samples to improve generalization when labeled data are scarce.

Let $\mathcal{T} = \mathcal{T}_l \cup \mathcal{T}_u$ denote a dataset consisting of a labeled set $\mathcal{T}_l = \{(\mathbf{x}_i, y_i)\}_{i=1}^{N_l}$ and an label-free set $\mathcal{T}_u = \{\mathbf{x}_i\}_{i=1}^{N_u}$. The goal of coreset selection in SSL is to construct a compact subset $\mathcal{S} \subset \mathcal{T}_u$ such that training on $\mathcal{T}_l \cup \mathcal{S}$ results in a model with a similar or superior generalization to training on the full dataset.

Formally, RETRIEVE optimizes a bilevel objective:

\begin{equation}
\mathcal{S}^* = \arg\min_{\mathcal{S} \subset \mathcal{T}_u, \frac{|\mathcal{S}|}{|\mathcal{T}_u|} \leq 1 - \alpha} \sum_{(\mathbf{x}, y) \in \mathcal{T}_l} \mathcal{L}(\mathbf{x}, y; \theta^*(\mathcal{S})),
\end{equation}
\noindent
where $\mathcal{L}(\mathbf{x}, y; \theta)$ is the supervised loss (e.g., cross-entropy), and the inner optimization defines the optimal model parameters $\theta^*$ as:

\begin{equation}
\theta^*(\mathcal{S}) = \arg\min_{\theta} \left( \sum_{(\mathbf{x}, y) \in \mathcal{T}_l} \mathcal{L}(\mathbf{x}, y; \theta) + \lambda \sum_{\mathbf{x} \in \mathcal{S}} \mathcal{L}_u(\mathbf{x}; \theta) \right).
\end{equation}
\noindent
Here, $\mathcal{L}_u(\mathbf{x}; \theta)$ represents the unsupervised loss applied to label-free samples, typically a consistency regularization term such as Mean Squared Error (MSE) between different augmentations.

Solving the bilevel optimization directly is computationally intractable. 
Instead, RETRIEVE employs a one-step gradient-based approximation to estimate the impact of selecting an label-free sample $\mathbf{x} \in \mathcal{T}_u$ on the labeled set’s loss:

\begin{equation}
s(\mathbf{x}) = \left\| \nabla_\theta \sum_{(\mathbf{x}, y) \in \mathcal{T}_l} \mathcal{L}(\mathbf{x}, y; \theta) \right\|_2 - \lambda \left\| \nabla_\theta \mathcal{L}_u(\mathbf{x}; \theta) \right\|_2.
\end{equation}

The coreset $\mathcal{S}$ is then selected by ranking samples according to $s(\mathbf{x})$, ensuring that label-free samples with high influence on the labeled loss are prioritized.

RETRIEVE is explicitly designed for SSL, leveraging partial labels, and may be too impractical in fully labeled scenarios.

\subsubsection{GLISTER}
Another intriguing bilevel optimization method is GLISTER (GeneraLIzation-based data Subset selecTion for Efficient and Robust learning), which selects a subset that maximizes log-likelihood performance on a held-out validation set rather than just the training set \cite{killamsetty2021glister}. 
This shift prioritizes generalization while maintaining efficiency.

Mathematically, GLISTER optimizes as follows:
\begin{equation}
    \argmax_{\mathcal{S} \subseteq \mathcal{T} \setminus \mathcal{V}, \frac{|\mathcal{S}|}{|\mathcal{T} \setminus \mathcal{V}|} \leq 1 - \alpha} \mathcal{L}_\mathcal{V} \left( \mathcal{V}, \argmax_{\theta} \mathcal{L}_{\mathcal{T} \setminus \mathcal{V}}(\mathcal{S}, \theta) \right),
\end{equation}
where $\mathcal{V} \subset \mathcal{T}$ is the validation set, $\mathcal{L}_\mathcal{V}$ the validation loss, and $\mathcal{L}_{\mathcal{T} \setminus \mathcal{V}}$ is the training loss on the remaining data that are candidates for the coreset.

The key advantage of GLISTER is its focus on ``held-out'' performance that can improve real-world robustness (e.g., fewer overfitting concerns).
However, it requires setting aside a validation set $\mathcal{V}$, reducing the total data for training and, thereby, potentially removing beneficial samples.

\begin{figure}[t!]
    \begin{center}
        \includegraphics[width=\columnwidth]{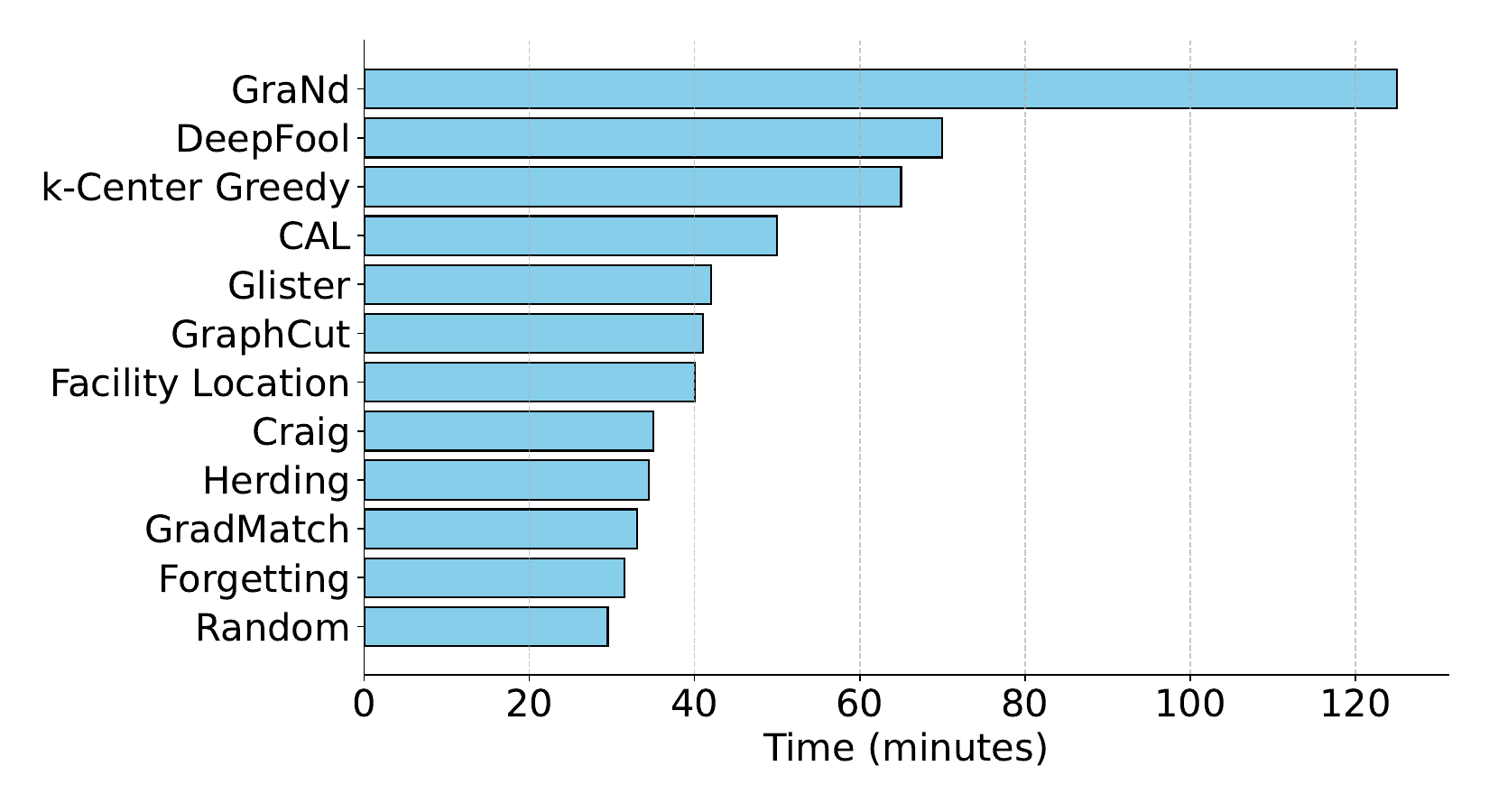}
        \caption{\label{fig:time}
        Comparison of method runtimes, selection plus training, in minutes (tested on RTX A6000), sorted in ascending order. The histogram shows the varying computational costs of the different selection strategies.
        }
    \end{center}
\end{figure}

\subsection{Computation Time}
\autoref{fig:time} illustrates the computation time for various coreset methods, sorted in ascending order. 
This comparison shows the trade-offs between computational efficiency and method complexity, which must be carefully considered when choosing a suitable approach for large-scale applications.
Notably, Random, Forgetting, GradMatch, Herding, and Craig exhibit the lowest computation times, with k-Center Greedy and DeepFool requiring significantly more time. The most computationally expensive method, GraNd, stands out with a runtime of 125 minutes (for ImageNet-1K, it takes 1.5 days on a single RTX A6000 GPU). 
Methods like CAL, Glister, GraphCut, and Facility Location fall in the mid-range, balancing computational cost and selection quality. 

\begin{table*}[!htbp]
\centering
\renewcommand{\arraystretch}{1.0}
\setlength{\tabcolsep}{4pt}
\caption{\label{tab:overview}Overview of coreset selection methods discussed in this review.}
\begin{tabular}{|p{0.1\textwidth}|p{0.2\textwidth}|p{0.14\textwidth}|p{0.5\textwidth}|}
\hline
\textbf{Category} & \textbf{Subcategory} & \textbf{Method} & \textbf{Comments} \\
\hline
\multirow{3}{*}{Training-free} 
    & Random & Random Selection &
      \begin{itemize}[noitemsep,topsep=0pt,parsep=0pt,partopsep=0pt]
          \item Simple and fast.
          \item Poor performance at high pruning rates.
      \end{itemize} \\ \cline{2-4}
    & Geometry-based & K-Center Greedy \cite{sener2017active} &
      \begin{itemize}[noitemsep,topsep=0pt,parsep=0pt,partopsep=0pt]
          \item Finds maximal coverage under a geometric measure.
          \item Suitable for clustering tasks.
          \item Computationally expensive for larger datasets.
      \end{itemize} \\ \cline{2-4}
    & Geometry-based & Herding \cite{welling2009herding} &
      \begin{itemize}[noitemsep,topsep=0pt,parsep=0pt,partopsep=0pt]
          \item Greedily selects samples to minimize distance between coreset and dataset centroids.
          \item Fast and easy to implement.
          \item Poor performance on complex data distributions.
      \end{itemize} \\
\hline
\multirow{10}{*}{Training-based} 
    & \multirow{2}{*}{Score-based} & Forgetting \cite{toneva2018empirical} &
      \begin{itemize}[noitemsep,topsep=0pt,parsep=0pt,partopsep=0pt]
          \item Counts forgetting events during training and 
picks samples with highest counts.
          \item Easy to implement.
          \item Highly non-deterministic.
      \end{itemize} \\ \cline{3-4}
    &  & GraNd \cite{paul2021deep} &
      \begin{itemize}[noitemsep,topsep=0pt,parsep=0pt,partopsep=0pt]
          \item Picks samples with maximal contribution to the loss function during training.
          \item Easy implementation.
          \item Expensive and slow for large datasets.
      \end{itemize} \\ \cline{2-4}
    & \multirow{2}{*}{Decision Boundary-based} & DeepFool \cite{ducoffe2018adversarial} &
      \begin{itemize}[noitemsep,topsep=0pt,parsep=0pt,partopsep=0pt]
          \item Iteratively finds the smallest perturbation to move samples across a decision boundary.
          \item Picks samples with smallest values.
          \item Expensive owing to repeated adversarial steps.
      \end{itemize} \\ \cline{3-4}
    &  & CAL \cite{margatina2021active} &
      \begin{itemize}[noitemsep,topsep=0pt,parsep=0pt,partopsep=0pt]
          \item Balances uncertainty and diversity by picking contrastive samples.
          \item Computationally intensive; Dependent on embedding quality.
      \end{itemize} \\ \cline{2-4}
    & \multirow{3}{*}{Submodularity-based} & FASS \cite{wei2015submodularity} &
      \begin{itemize}[noitemsep,topsep=0pt,parsep=0pt,partopsep=0pt]
          \item Combines informativeness and representativeness.
          \item Filters out uncertain samples.
          \item Computationally expensive.
      \end{itemize} \\ \cline{3-4}
    &  & PRISM \cite{kothawade2022prism} &
      \begin{itemize}[noitemsep,topsep=0pt,parsep=0pt,partopsep=0pt]
          \item Adaptable to different information measures; robust to noise.
          \item Requires careful parameter tuning.
          \item May not be efficient for very large datasets.
      \end{itemize} \\ \cline{3-4}
    &  & SIMILAR \cite{kothawade2021similar} &
      \begin{itemize}[noitemsep,topsep=0pt,parsep=0pt,partopsep=0pt]
          \item Captures both informativeness and diversity.
          \item Computationally expensive due to submodular optimization.
          \item Requires labeled data.
      \end{itemize} \\ \cline{2-4}
    & \multirow{2}{*}{Gradient Matching-based} & CRAIG \cite{mirzasoleiman2020coresets} &
      \begin{itemize}[noitemsep,topsep=0pt,parsep=0pt,partopsep=0pt]
          \item Efficiently accelerates gradient descent.
          \item Reduces computational cost.
          \item Limited to specific optimization algorithms.
          \item May not generalize well to all tasks.
      \end{itemize} \\ \cline{3-4}
    &  & Gradmatch \cite{killamsetty2021grad} &
      \begin{itemize}[noitemsep,topsep=0pt,parsep=0pt,partopsep=0pt]
          \item Matches gradients of the coreset to the full dataset.
          \item Effective for neural networks.
          \item Requires computing gradients for the entire dataset.
          \item Sensitive to hyperparameters.
      \end{itemize} \\ \cline{2-4}
    & \multirow{2}{*}{Bilevel Optimization-based} & RETRIEVE \cite{killamsetty2021retrieve} &
      \begin{itemize}[noitemsep,topsep=0pt,parsep=0pt,partopsep=0pt]
          \item Iteratively refines coreset based on model predictions.
          \item Computationally intensive due to iterative refinement.
          \item Requires model retraining.
      \end{itemize} \\ \cline{3-4}
    &  & GLISTER \cite{killamsetty2021glister} &
      \begin{itemize}[noitemsep,topsep=0pt,parsep=0pt,partopsep=0pt]
          \item Generalizes well across different models; Robust to noise and outliers.
          \item Requires careful tuning of generalization metrics.
          \item May not be efficient for very large datasets.
      \end{itemize} \\
\hline
\end{tabular}
\label{tab:methods-overview}
\end{table*}

\begin{table}[t!]
\centering
\caption{Hyperparameter sensitivity of representative coreset selectors.
We list common sensitive knobs and typical failure modes when mis-set.}
\label{tab:hparam_sensitivity}
\renewcommand{\arraystretch}{1.10}
\setlength{\tabcolsep}{3pt}
\scriptsize
\begin{tabularx}{\columnwidth}{@{} p{0.17\columnwidth} X X @{}}
\toprule
\textbf{Method} & \textbf{Sensitive knobs (examples)} & \textbf{Typical failure mode} \\
\midrule
\textbf{GraNd/EL2N} &
Checkpoint/epoch for scoring; scoring frequency; gradient batch size; seed averaging &
Early: noisy/hard bias. Late: score collapse or outlier dominance. High variance across seeds. \\

\textbf{CAL} &
Encoder/embedding quality; neighborhood size $k$; similarity threshold $\epsilon$; KL/uncertainty weighting; temperature &
Near-duplicate selection (too strict) or drift to pure uncertainty sampling (too weak similarity constraint); unstable under embedding/domain shift. \\

\textbf{GradMatch} &
OMP iterations/budget; regularization $\lambda$; gradient refresh frequency; layer choice; mini- vs full-batch gradients &
Myopic greedy picks and missed coverage; noisy/oscillatory selections under gradient noise; high runtime if gradients are recomputed too often. \\

\textbf{Submodular} &
Similarity metric; kernel bandwidth/temperature; candidate pool size; lazy-greedy settings; coverage vs diversity weighting &
Kernel miscalibration yields duplicates (too peaky) or overspread subsets (too flat); memory/time blow-up from dense similarity matrices. \\

\textbf{DeepFool} &
Max iterations; overshoot/step size; norm choice ($\ell_2$ vs $\ell_\infty$); stopping criterion; initialization &
Non-convergence (underestimates boundary proximity) or overshoot (breaks ranking); high compute and sensitivity to calibration. \\

\textbf{GLISTER} &
Validation split size/quality; proxy inner steps; proxy LR/regularization; selection batch size &
Overfits to small/biased validation; unstable gains when proxy training is too short; brittle subsets across seeds/architectures. \\
\bottomrule
\end{tabularx}
\vspace{-1em}
\end{table}

\section{Label-free Coreset Selection}
\label{sec:blind}
Traditional coreset selection methods rely on labeled data to estimate sample importance via loss values, gradient norms, or decision boundary distances. 
In contrast, label-free coreset selection methods aim to approximate these training dynamics without ground-truth annotations.

\begin{figure}[t!]
    \begin{center}
        \includegraphics[width=.85\columnwidth]{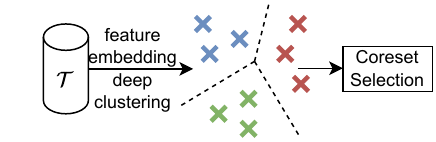}
        \caption{\label{fig:ELFS}
        Illustration of ELFS. First, it embeds the dataset in a feature space embedding and applies deep clustering to assign pseudo-labels. Next, it applies coreset selection based on scoring (forgetting, gradient norms, and sensitivity).
        }
    \end{center}
\end{figure}

\subsection{ELFS}
The Estimation of Label-Free Sample Scores (ELFS) framework \cite{zheng2024elfs} achieves label-free coreset selection by leveraging pseudo-labeling through deep clustering \cite{ren2024deep} to approximate informative selection criteria.
More formally, given an label-free dataset $\mathcal{T} = \{\mathbf{x}_i\}_{i=1}^{N}$, ELFS estimates pseudo-labels $\hat{y}_i$ by clustering feature embeddings extracted from a pre-trained network $\phi$:

\begin{equation}
\mathbf{z}_i = \phi(\mathbf{x}_i) \quad \forall \mathbf{x}_i \in \mathcal{T}.
\end{equation}
The pseudo-label assignment is then performed via deep clustering:
\begin{equation}
\hat{y}_i = \arg\max_k p(y = k | \mathbf{z}_i),
\end{equation}
\noindent
where $p(y | \mathbf{z}_i)$ is estimated using a clustering algorithm such as teacher ensemble-weighted mutual information \cite{adaloglou2023exploring}. 
Once pseudo-labels are assigned, ELFS estimates training dynamics metrics, such as forgetting events, gradient norms, and sample sensitivity.

The whole procedure is visualized in \autoref{fig:ELFS}, namely first applying deep clustering and then classical coreset selection methods afterward.
By leveraging pseudo-labels, it mimics training-informed, well-known coreset selection methods (e.g., forgetting events, gradient norms) without explicit supervision.
However, the effectiveness of ELFS is highly dependent on how well the clustering algorithm assigns pseudo-labels.
This problem becomes pronounced in high-dimensional embeddings, where feature separability is challenging, leading to poor clustering performance.

\begin{figure*}[th!]
    \begin{center}
        \includegraphics[width=.7\textwidth]{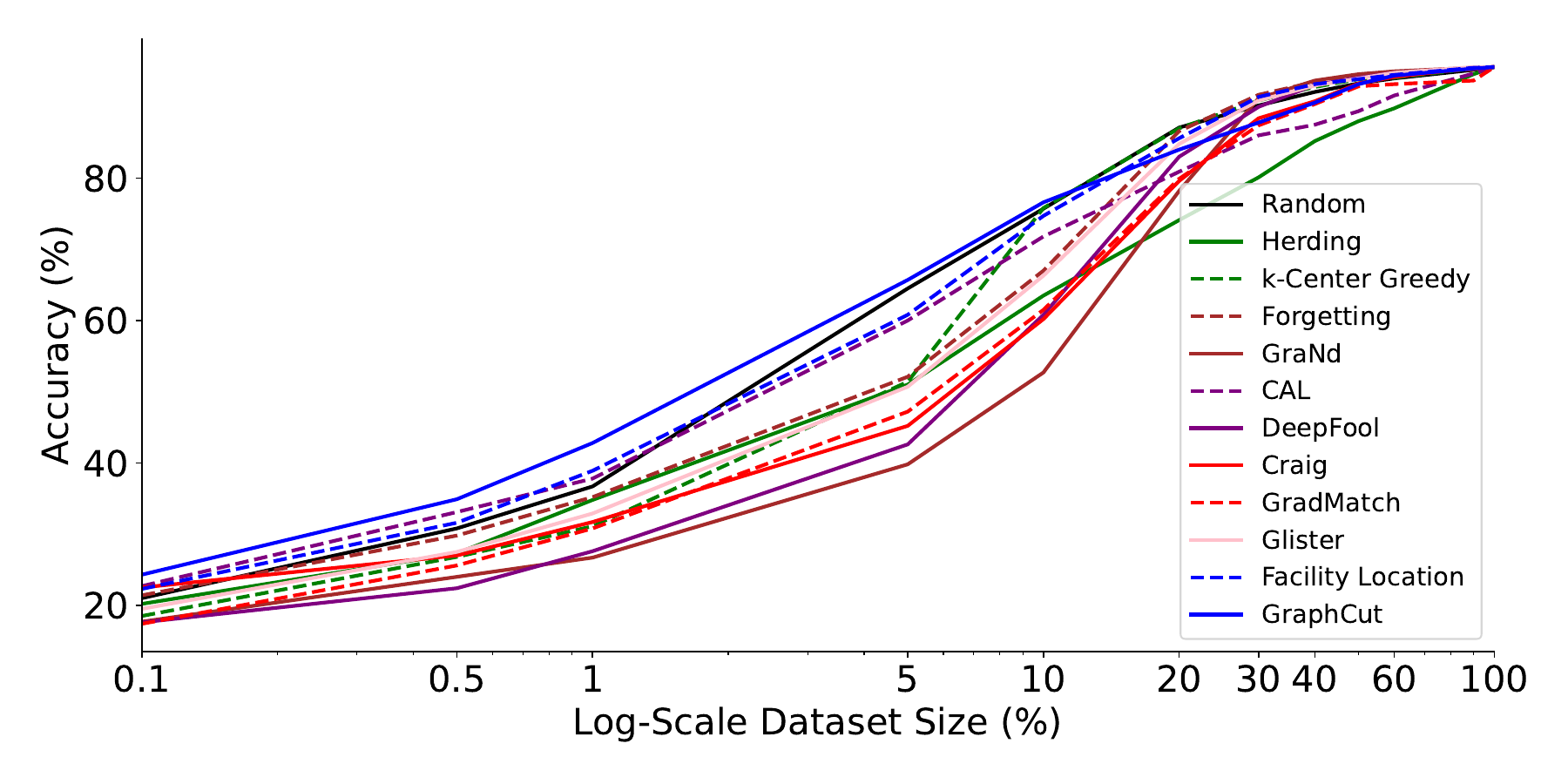}
        \caption{\label{fig:cifar10}
        Coreset selection performances on CIFAR-10. Methods grouped by our taxonomy a colored the same, with dashed lines indicating the different methods within the same category. 
        A general observation is that methods that are good for high pruning ratios ($\alpha$) fail to maintain the performance at low pruning ratios and vice versa. We can see a transition between top-performing methods between 5\% and 30\% of the remaining dataset size. Concrete values can be found in \autoref{tab:cifar}.
        Values are transcribed from DeepCore \cite{guo2022deepcore}.
        }
    \end{center}
\end{figure*}

\subsection{ZCore}
Zero-shot Core (ZCore) circumvents the need for model-specific supervision or clustering by employing pre-trained Vision-Language Models (VLMs) to encode feature representations \cite{griffin2024zero}. 
Given an label-free dataset $\mathcal{T}$, the method first embeds each sample into a feature space using a pre-trained foundation model $\phi$ (similar to geometry-based methods):
\begin{equation}
    \mathbf{z}_i = \phi(\mathbf{x}_i) \quad \forall \mathbf{x}_i \in \mathcal{T}
\end{equation}

Next, ZCore applies a coverage-based selection strategy inspired by k-Center Greedy and submodular maximization. 
The goal is to identify a subset such that every point in $\mathcal{T}$ is close to at least one representative sample in $\mathcal{S}$, which is expressed as a score function $s$:
\begin{equation}
    \mathcal{S} = \argmax_{\mathcal{S} \subset \mathcal{T}, \frac{|\mathcal{S}|}{|\mathcal{T}|} \leq 1 - \alpha} \sum_{\mathbf{x}_i \in \mathcal{S}} s(\mathbf{x}_i).
\end{equation}

The key advantage of ZCore is that by eliminating the dependence on model-specific training, it can be applied to any label-free dataset without additional computational overhead.
However, its reliance on pre-trained representations introduces potential domain bias (similar to other training-free methods).

Additionally, while ZCore ensures geometric coverage, it does not explicitly account for task-specific importance, which may limit performance in highly specialized applications.

\subsection{Other Notable Mentions}
Some very recent works attempt to combine diversity with difficulty without labels. 
For example, D2Pruning (Diversity \& Difficulty Pruning) \cite{maharana2023d2} constructs a graph of data points (nodes) with edges linking similar samples, then spreads “difficulty” scores across this graph to select a subset of examples that are both diverse and hard.
They report improved performance over prior methods at moderate pruning rates by capturing complementary aspects of data utility. Another example is SimCore \cite{kim2023coreset}, which targets an open-set self-supervised learning scenario by selecting from a large unlabeled pool those samples most similar to a given target dataset’s domain. 
While SimCore is slightly tangential (aimed at data augmentation for representation learning), it reinforces the trend of using semantic similarity in feature space.

\section{Theoretical Remarks}
\label{sec:theoretical-remarks}
A fundamental question in coreset selection is whether a carefully chosen subset $\mathcal{S}$ can match or even surpass the performance of the full dataset $\mathcal{T}$. 
This question also sits within the broader area of dataset pruning.
This challenge connects to neural scaling laws \cite{kaplan2020scaling, hestness2017deep, cherti2023reproducible,gao2023scaling}. 
These laws suggest that the generalization error $\epsilon(\mathcal{T})$ typically follows a power-law decay:
\begin{equation}
    \epsilon(\mathcal{T}) \propto |\mathcal{T}|^{-\beta}
\end{equation}
\noindent
where $\beta$ is a dataset-dependent exponent. 
This suggests that collecting more data improves model performance but with diminishing returns. 
However, coreset selection (and pruning more broadly) can disrupt this traditional scaling. 
The reason is selection bias: pruning favors the most informative samples \cite{isik2022information, vysogorets2025drop}. 
As a result, a well-pruned dataset may deviate from the standard power law:
\begin{equation}
    \label{eq:better}
    \epsilon(\mathcal{S}) \propto |\mathcal{S}|^{-\beta{\prime}}, \quad \text{ where } \beta{\prime} > \beta.
\end{equation}

Thus, the right selection can generalize better than a random subset of the same size.
The effective information content of a dataset can be defined as:
\begin{equation}
    I_{\text{eff}}(\mathcal{T}) = \sum_{(\mathbf{x}, y) \in \mathcal{T}} w(\mathbf{x}, y) H(\mathbf{x}, y),
\end{equation}
\noindent
where $w(\mathbf{x}, y)$ is a weighting function that adjusts for redundancy, and $H(\mathbf{x}, y)$ is the sample entropy \cite{richman2000physiological}. 
By filtering out samples with the lowest information contribution, ensuring that only the most representative data remains.

\textit{Sorscher et al. }\cite{sorscher2022beyond} provide a deeper analysis of how pruning affects generalization. 
They argue that the optimal subset selection strategy depends on the size of the original dataset:
\begin{itemize}
    \item For small datasets ($|\mathcal{T}|$ is small), keeping easy (high-margin) samples is better. These samples provide coarse-grained information, which prevents overfitting and ensures stable learning.
    \item For large datasets ($|\mathcal{T}|$ is large), keeping hard (low-margin) samples is more effective. These samples refine the decision boundary and contribute to fine-grained generalization.
\end{itemize}
They further argue that the pruning ratio should increase with dataset size. 
In large datasets, much of the information is redundant, so a smaller fraction of harder examples can suffice.
We observe similar trends on CIFAR-10 (see \autoref{fig:cifar10} and \autoref{tab:cifar}). 
These results are consistent with the above insights. 
Easy-sample strategies perform better at high pruning ratios. 

They often underperform at lower ratios, and the reverse holds for hard-sample strategies. 
We observe take-over points around a 10\% dataset size.
However, \textit{Zheng et al. }\cite{zheng2022coverage} and others \cite{xia2022moderate} partially disagree. 
They report that retaining a small fraction of redundant samples can further improve performance when combined with hardest-first sampling.

\begin{table*}[t!]

  \centering
  \caption{Coreset selection performances on CIFAR10 with randomly initialized ResNet-18 \cite{he2016deep} models. Values are transcribed from DeepCore \cite{guo2022deepcore}.}  \label{tab:cifar}
  \resizebox{\linewidth}{!}{
 \footnotesize{
 \setlength{\tabcolsep}{2pt}
\begin{tabular}{ccccccccccccc}
\toprule
Fraction $(1-\alpha)$ & 0.1\%       & 0.5\%       & 1\%       & 5\%       & 10\%      & 20\%      & 30\%      & 40\%      & 50\%      & 60\%      & 90\%      & 100\%        \\ \cmidrule(lr){1-13}
Random         & 21.0\hspace{0.02em}$\pm$\hspace{0.02em}0.3          & 30.8\hspace{0.02em}$\pm$\hspace{0.02em}0.6          & 36.7\hspace{0.02em}$\pm$\hspace{0.02em}1.7          & 64.5\hspace{0.02em}$\pm$\hspace{0.02em}1.1          & 75.7\hspace{0.02em}$\pm$\hspace{0.02em}2.0          & \textbf{87.1\hspace{0.02em}$\pm$\hspace{0.02em}0.5}          & 90.2\hspace{0.02em}$\pm$\hspace{0.02em}0.3          & 92.1\hspace{0.02em}$\pm$\hspace{0.02em}0.1          & 93.3\hspace{0.02em}$\pm$\hspace{0.02em}0.2          & 94.0\hspace{0.02em}$\pm$\hspace{0.02em}0.2          & 95.2\hspace{0.02em}$\pm$\hspace{0.02em}0.1          & 95.6\hspace{0.02em}$\pm$\hspace{0.02em}0.1 \\ \cmidrule(lr){1-13}
Herding   \cite{welling2009herding}  & 20.2\hspace{0.02em}$\pm$\hspace{0.02em}2.3          & 27.3\hspace{0.02em}$\pm$\hspace{0.02em}1.5          & 34.8\hspace{0.02em}$\pm$\hspace{0.02em}3.3          & 51.0\hspace{0.02em}$\pm$\hspace{0.02em}3.1          & 63.5\hspace{0.02em}$\pm$\hspace{0.02em}3.4          & 74.1\hspace{0.02em}$\pm$\hspace{0.02em}2.5          & 80.1\hspace{0.02em}$\pm$\hspace{0.02em}2.2          & 85.2\hspace{0.02em}$\pm$\hspace{0.02em}0.9          & 88.0\hspace{0.02em}$\pm$\hspace{0.02em}1.1          & 89.8\hspace{0.02em}$\pm$\hspace{0.02em}0.9          & 94.6\hspace{0.02em}$\pm$\hspace{0.02em}0.4          & 95.6\hspace{0.02em}$\pm$\hspace{0.02em}0.1 \\
k-Center Greedy \cite{sener2017active} & 18.5\hspace{0.02em}$\pm$\hspace{0.02em}0.3          & 26.8\hspace{0.02em}$\pm$\hspace{0.02em}1.2          & 31.1\hspace{0.02em}$\pm$\hspace{0.02em}1.2          & 51.4\hspace{0.02em}$\pm$\hspace{0.02em}2.1          & 75.8\hspace{0.02em}$\pm$\hspace{0.02em}2.4          & 87.0\hspace{0.02em}$\pm$\hspace{0.02em}0.3          & 90.9\hspace{0.02em}$\pm$\hspace{0.02em}0.4          & 92.8\hspace{0.02em}$\pm$\hspace{0.02em}0.1          & 93.9\hspace{0.02em}$\pm$\hspace{0.02em}0.2          & 94.1\hspace{0.02em}$\pm$\hspace{0.02em}0.1          & 95.4\hspace{0.02em}$\pm$\hspace{0.02em}0.1          & 95.6\hspace{0.02em}$\pm$\hspace{0.02em}0.1 \\ \cmidrule(lr){1-13}
Forgetting   \cite{toneva2018empirical}  & 21.4\hspace{0.02em}$\pm$\hspace{0.02em}0.5 & 29.8\hspace{0.02em}$\pm$\hspace{0.02em}1.0 & 35.2\hspace{0.02em}$\pm$\hspace{0.02em}1.6 & 52.1\hspace{0.02em}$\pm$\hspace{0.02em}2.2 & 67.0\hspace{0.02em}$\pm$\hspace{0.02em}1.5 & 86.6\hspace{0.02em}$\pm$\hspace{0.02em}0.6 & \textbf{91.7\hspace{0.02em}$\pm$\hspace{0.02em}0.3} & 93.5\hspace{0.02em}$\pm$\hspace{0.02em}0.2 & 94.1\hspace{0.02em}$\pm$\hspace{0.02em}0.1 & 94.6\hspace{0.02em}$\pm$\hspace{0.02em}0.2 & 95.3\hspace{0.02em}$\pm$\hspace{0.02em}0.1& 95.6\hspace{0.02em}$\pm$\hspace{0.02em}0.1 \\
GraNd  \cite{paul2021deep}   & 17.7\hspace{0.02em}$\pm$\hspace{0.02em}1.0 & 24.0\hspace{0.02em}$\pm$\hspace{0.02em}1.1 & 26.7\hspace{0.02em}$\pm$\hspace{0.02em}1.3 & 39.8\hspace{0.02em}$\pm$\hspace{0.02em}2.3 & 52.7\hspace{0.02em}$\pm$\hspace{0.02em}1.9 & 78.2\hspace{0.02em}$\pm$\hspace{0.02em}2.9 & 91.2\hspace{0.02em}$\pm$\hspace{0.02em}0.7 & \textbf{93.7\hspace{0.02em}$\pm$\hspace{0.02em}0.3} & \textbf{94.6\hspace{0.02em}$\pm$\hspace{0.02em}0.1} & \textbf{95.0\hspace{0.02em}$\pm$\hspace{0.02em}0.2} & 95.5\hspace{0.02em}$\pm$\hspace{0.02em}0.2          & 95.6\hspace{0.02em}$\pm$\hspace{0.02em}0.1 \\ \cmidrule(lr){1-13}
CAL   \cite{margatina2021active}      & 22.7\hspace{0.02em}$\pm$\hspace{0.02em}2.7          & 33.1\hspace{0.02em}$\pm$\hspace{0.02em}2.3          & 37.8\hspace{0.02em}$\pm$\hspace{0.02em}2.0          & 60.0\hspace{0.02em}$\pm$\hspace{0.02em}1.4          & 71.8\hspace{0.02em}$\pm$\hspace{0.02em}1.0          & 80.9\hspace{0.02em}$\pm$\hspace{0.02em}1.1          & 86.0\hspace{0.02em}$\pm$\hspace{0.02em}1.9          & 87.5\hspace{0.02em}$\pm$\hspace{0.02em}0.8          & 89.4\hspace{0.02em}$\pm$\hspace{0.02em}0.6          & 91.6\hspace{0.02em}$\pm$\hspace{0.02em}0.9          & 94.7\hspace{0.02em}$\pm$\hspace{0.02em}0.3          & 95.6\hspace{0.02em}$\pm$\hspace{0.02em}0.1 \\
DeepFool     \cite{ducoffe2018adversarial}    & 17.6\hspace{0.02em}$\pm$\hspace{0.02em}0.4          & 22.4\hspace{0.02em}$\pm$\hspace{0.02em}0.8          & 27.6\hspace{0.02em}$\pm$\hspace{0.02em}2.2          & 42.6\hspace{0.02em}$\pm$\hspace{0.02em}3.5          & 60.8\hspace{0.02em}$\pm$\hspace{0.02em}2.5          & 83.0\hspace{0.02em}$\pm$\hspace{0.02em}2.3          & 90.0\hspace{0.02em}$\pm$\hspace{0.02em}0.7          & 93.1\hspace{0.02em}$\pm$\hspace{0.02em}0.2          & 94.1\hspace{0.02em}$\pm$\hspace{0.02em}0.1          & 94.8\hspace{0.02em}$\pm$\hspace{0.02em}0.2          & 95.5\hspace{0.02em}$\pm$\hspace{0.02em}0.1          & 95.6\hspace{0.02em}$\pm$\hspace{0.02em}0.1 \\ \cmidrule(lr){1-13}
Craig     \cite{mirzasoleiman2020coresets}   & 22.5\hspace{0.02em}$\pm$\hspace{0.02em}1.2          & 27.0\hspace{0.02em}$\pm$\hspace{0.02em}0.7          & 31.7\hspace{0.02em}$\pm$\hspace{0.02em}1.1          & 45.2\hspace{0.02em}$\pm$\hspace{0.02em}2.9          & 60.2\hspace{0.02em}$\pm$\hspace{0.02em}4.4          & 79.6\hspace{0.02em}$\pm$\hspace{0.02em}3.1          & 88.4\hspace{0.02em}$\pm$\hspace{0.02em}0.5          & 90.8\hspace{0.02em}$\pm$\hspace{0.02em}1.4          & 93.3\hspace{0.02em}$\pm$\hspace{0.02em}0.6          & 94.2\hspace{0.02em}$\pm$\hspace{0.02em}0.2          & 95.5\hspace{0.02em}$\pm$\hspace{0.02em}0.1          & 95.6\hspace{0.02em}$\pm$\hspace{0.02em}0.1 \\
GradMatch   \cite{killamsetty2021grad}  & 17.4\hspace{0.02em}$\pm$\hspace{0.02em}1.7          & 25.6\hspace{0.02em}$\pm$\hspace{0.02em}2.6          & 30.8\hspace{0.02em}$\pm$\hspace{0.02em}1.0          & 47.2\hspace{0.02em}$\pm$\hspace{0.02em}0.7          & 61.5\hspace{0.02em}$\pm$\hspace{0.02em}2.4          & 79.9\hspace{0.02em}$\pm$\hspace{0.02em}2.6          & 87.4\hspace{0.02em}$\pm$\hspace{0.02em}2.0          & 90.4\hspace{0.02em}$\pm$\hspace{0.02em}1.5          & 92.9\hspace{0.02em}$\pm$\hspace{0.02em}0.6          & 93.2\hspace{0.02em}$\pm$\hspace{0.02em}1.0          & 93.7\hspace{0.02em}$\pm$\hspace{0.02em}0.5          & 95.6\hspace{0.02em}$\pm$\hspace{0.02em}0.1 \\ \cmidrule(lr){1-13}
Glister  \cite{killamsetty2021glister}  & 19.5\hspace{0.02em}$\pm$\hspace{0.02em}2.1          & 27.5\hspace{0.02em}$\pm$\hspace{0.02em}1.4          & 32.9\hspace{0.02em}$\pm$\hspace{0.02em}2.4          & 50.7\hspace{0.02em}$\pm$\hspace{0.02em}1.5          & 66.3\hspace{0.02em}$\pm$\hspace{0.02em}3.5          & 84.8\hspace{0.02em}$\pm$\hspace{0.02em}0.9          & 90.9\hspace{0.02em}$\pm$\hspace{0.02em}0.3          & 93.0\hspace{0.02em}$\pm$\hspace{0.02em}0.2          & 94.0\hspace{0.02em}$\pm$\hspace{0.02em}0.3          & 94.8\hspace{0.02em}$\pm$\hspace{0.02em}0.2          & \textbf{95.6\hspace{0.02em}$\pm$\hspace{0.02em}0.2} & 95.6\hspace{0.02em}$\pm$\hspace{0.02em}0.1 \\ \cmidrule(lr){1-13}
Facility Location     \cite{iyer2021submodular}    & 22.3\hspace{0.02em}$\pm$\hspace{0.02em}2.0          & 31.6\hspace{0.02em}$\pm$\hspace{0.02em}0.6          & 38.9\hspace{0.02em}$\pm$\hspace{0.02em}1.4          & 60.8\hspace{0.02em}$\pm$\hspace{0.02em}2.5          & 74.7\hspace{0.02em}$\pm$\hspace{0.02em}1.3          & 85.6\hspace{0.02em}$\pm$\hspace{0.02em}1.9          & 91.4\hspace{0.02em}$\pm$\hspace{0.02em}0.4 & 93.2\hspace{0.02em}$\pm$\hspace{0.02em}0.3          & 93.9\hspace{0.02em}$\pm$\hspace{0.02em}0.2          & 94.5\hspace{0.02em}$\pm$\hspace{0.02em}0.3          & 95.5\hspace{0.02em}$\pm$\hspace{0.02em}0.2          & 95.6\hspace{0.02em}$\pm$\hspace{0.02em}0.1 \\
GraphCut   \cite{iyer2021submodular}    &   \textbf{24.3\hspace{0.02em}$\pm$\hspace{0.02em}1.5} & \textbf{34.9\hspace{0.02em}$\pm$\hspace{0.02em}2.3} & \textbf{42.8\hspace{0.02em}$\pm$\hspace{0.02em}1.3} & \textbf{65.7\hspace{0.02em}$\pm$\hspace{0.02em}1.2} & \textbf{76.6\hspace{0.02em}$\pm$\hspace{0.02em}1.5} & 84.0\hspace{0.02em}$\pm$\hspace{0.02em}0.5          & 87.8\hspace{0.02em}$\pm$\hspace{0.02em}0.4          & 90.6\hspace{0.02em}$\pm$\hspace{0.02em}0.3          & 93.2\hspace{0.02em}$\pm$\hspace{0.02em}0.3          & 94.4\hspace{0.02em}$\pm$\hspace{0.02em}0.3          & 95.4\hspace{0.02em}$\pm$\hspace{0.02em}0.1          & 95.6\hspace{0.02em}$\pm$\hspace{0.02em}0.1\\  
\bottomrule
\end{tabular}
}
}
\end{table*}

\begin{figure*}[t!]
    \begin{center}
        \includegraphics[width=.65\textwidth]{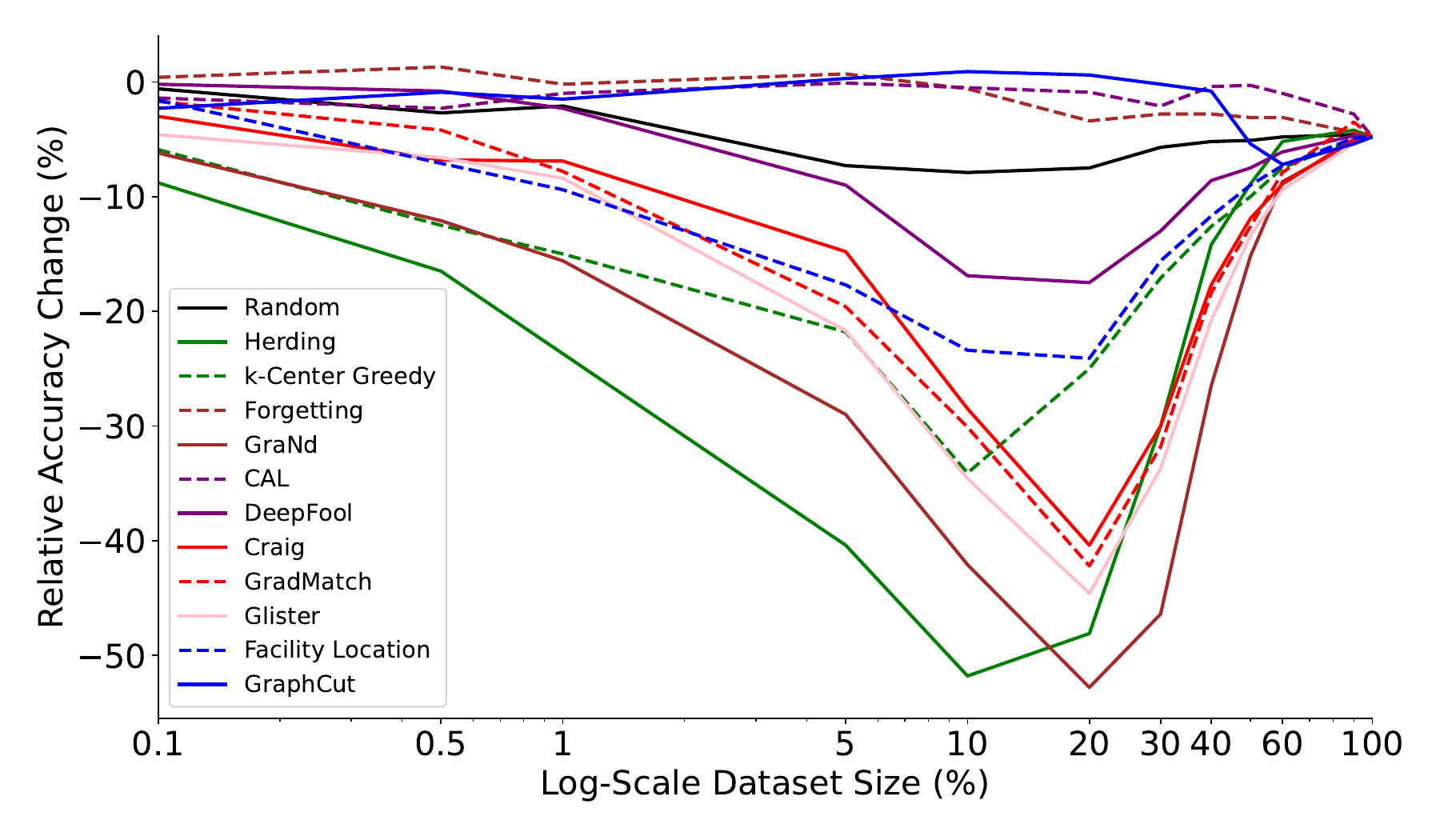}
        \caption{\label{fig:cifar10_10}
        Coreset selection performances on CIFAR-10 with 10\% corrupted labels, which means that 10\% of samples have been reassigned to new class labels \cite{zhang2021understanding}. 
        Methods are grouped by our taxonomy and share the same color. 
        Dashed lines distinguish methods within the same category. 
        Most methods show a sharp drop at mid-range pruning ratios (10\%--50\%). 
        This is most pronounced for geometry-based methods, gradient matching approaches, and GraNd (gradient norm).
        Forgetting, CAL, and GraphCut remain more stable in comparison to the other methods. 
        Concrete values can be found in \autoref{tab:corrupted}.
        Methods were evaluated based on DeepCore \cite{guo2022deepcore} under identical settings.
        }
    \end{center}
\end{figure*}

A key takeaway is that pruning can break traditional power-law scaling.  
While random subsampling follows the expected decay $\beta$, a well-pruned dataset can achieve an improved scaling exponent $\beta{\prime}$, leading to significant improvements. 

\subsection{Implications for Model Design and Loss Construction}
Theoretical discussions of pruning often stop at error-vs.-data scaling (e.g., \autoref{eq:better}), but in practice coreset selection also changes the \emph{training signal geometry}: it modifies (i) the distribution of margins/difficulties, (ii) the gradient-noise statistics, and (iii) the degree of redundancy. These shifts have concrete implications for model capacity, loss choices, and optimization hyperparameters.

\paragraph{Model design}
A useful lens is that pruning changes the dataset's \emph{effective sample size} (ESS), which depends on how non-uniform the subset is.
If samples in a coreset are used with weights $\{w_i\}_{i\in\mathcal{S}}$, a standard proxy for statistical efficiency is
\begin{equation}
\mathrm{ESS}(\mathcal{S}) \;\triangleq\; \frac{\big(\sum_{i\in\mathcal{S}} w_i\big)^2}{\sum_{i\in\mathcal{S}} w_i^2}\;\;\le\;\;|\mathcal{S}|.
\end{equation}

Highly skewed selection (e.g., hardest-first scoring without smoothing) can yield $\mathrm{ESS}\ll|\mathcal{S}|$, meaning the model effectively trains on fewer independent samples.
This predicts a practical failure mode: \emph{large-capacity models overfit more easily under aggressive, high-variance pruning}.
Quantitatively, if the original training setup was near the interpolation/over-parameterized regime, a conservative guideline is to keep the ratio $\frac{P}{\mathrm{ESS}}$ (parameters per effective sample) from increasing drastically when moving from $\mathcal{T}$ to $\mathcal{S}$.
When $\mathrm{ESS}$ collapses, one may need stronger regularization (weight decay, augmentation, dropout) or smaller models to maintain generalization.

\paragraph{Loss Construction}
Coreset objectives implicitly reshape the distribution of margins and label noise:
hard-sample selectors (GraNd/EL2N, boundary-based) bias toward low-margin, high-loss points,
while coverage/submodular selectors bias toward distributional support.
This matters because standard cross-entropy is sensitive to mislabeled outliers.
Hence, under noisy or outlier-heavy pruning, robustification becomes more important.
Two practical, quantitative knobs are:
(i) \emph{label smoothing} (e.g., replace one-hot $y$ with $(1-\epsilon)y+\epsilon/K$),
and (ii) \emph{loss clipping / robust losses} that reduce the influence of extreme-loss samples.

\subsection{Concerning Other Modalities}

While coreset selection is applicable across modalities (see \autoref{sec:applications}), most \emph{comparable} empirical evidence is concentrated in image classification benchmarks, where selection pipelines, architectures, and evaluation protocols are relatively standardized (e.g., fixed training recipes, shared backbones, and consistent metrics), enabling controlled cross-method comparisons.
In contrast, large-scale vision (e.g., ImageNet-scale) and NLP pretraining introduce confounds (tokenization and sequence length effects, heavy deduplication/filtering pipelines, domain-mixing, and substantial training-budget sensitivity) that make “apples-to-apples” benchmarking across coreset objectives considerably harder and often entangled with data quality filtering rather than subset selection alone.

\subsection{Outliers}
Another interesting aspect of coreset selection, often overlooked, is the role of outliers \cite{ruff2018deep, anwar2024fedad}. 
Outliers can either provide valuable decision boundary information or introduce noise and perturbations during the training. 
The prior handling of outliers, e.g., with deep SVDD \cite{ruff2018deep,zhou2021vae} or contrastive score-based OOD \cite{wang2025synco}, can significantly improve the performance of any coreset selection method \cite{wang2021robust}. 

A practical caveat is that, in many real-world settings, the presence (and rate) of outliers or label noise is \emph{unknown} a priori.
Most coreset selection methods are developed and tuned under an implicit \emph{clean-data assumption}.
When this assumption is violated, selection scores can become dominated by mislabeled or anomalous samples, leading to highly skewed subsets and drastic performance drops.
This motivates treating outlier/OOD filtering not as an optional pre-processing step, but as a robustness-critical component of any future coreset pipeline, especially under aggressive pruning.

We provide in \autoref{fig:cifar10_10} and \autoref{tab:corrupted} additional robustness experiments to highlight the importance of outliers. 
In this experiment, we corrupted 10\% of the labels, i.e., reassigning 10\% of samples to new class labels \cite{zhang2021understanding}. 
The selection of corrupted labels was repeated during each coreset sampling procedure.
To make the empirical comparisons interpretable, we emphasize that the results follow an identical training/evaluation protocol across methods. 
Concretely, we evaluated a ResNet-18 with SGD (initial learning rate $0.1$ with cosine decay, momentum $0.9$, weight decay $5\times10^{-4}$), trained for $200$ epochs with standard CIFAR-10 augmentation (random crop with $4$-pixel padding and random horizontal flip). 
Training-oriented selectors compute their scores from a partially trained network (a $10$-epoch pretraining stage for scoring), and reported numbers are averaged over $5$ random seeds (mean $\pm$ std), as provided by the benchmark.

As a result, one can see that most sampling methods get negatively affected (red), especially geometry-based, gradient-matching-based methods, and GraNd (gradient norm).
Interestingly, forgetting, CAL, and GraphCut remain more stable in comparison to the other methods.
Furthermore, CAL confirms \autoref{eq:better}, as its $\alpha=10\%$ pruned dataset outperforms the training on the full (corrupted) dataset.
Also, forgetting and GraphCut were able to achieve better results for some cases.

As this experiment highlights, future work could further explore automated ways prior to coreset selection to filter outliers dynamically.
For instance, \textit{Zheng et al.} \cite{zheng2022coverage} leveraged the area under margin \cite{pleiss2020identifying}, which captures how far a sample is from the decision boundary over multiple training iterations.
Other innovations might include classical outlier detection principles, as shown in HyperCore \cite{moser2025hypercore}.

\begin{table*}[t!]

  \centering
  \caption{Robustness on CIFAR10 with 10\% corrupted labels (i.e., 10\% of total samples, randomly selected, have been reassigned to new class labels). Methods were evaluated based on DeepCore \cite{guo2022deepcore} under identical settings.}  \label{tab:corrupted}
  \resizebox{\linewidth}{!}{
 \footnotesize{
 \setlength{\tabcolsep}{2pt}
\begin{tabular}{ccccccccccccc}
\toprule
Fraction $(1-\alpha)$ & 0.1\%       & 0.5\%       & 1\%       & 5\%       & 10\%      & 20\%      & 30\%      & 40\%      & 50\%      & 60\%      & 90\%      & 100\%        \\ \cmidrule(lr){1-13}
Random         & 
20.4\hspace{0.02em}$\pm$\hspace{0.02em}0.3          & 28.1\hspace{0.02em}$\pm$\hspace{0.02em}1.8          & 34.6\hspace{0.02em}$\pm$\hspace{0.02em}1.0          & 57.2\hspace{0.02em}$\pm$\hspace{0.02em}1.5          & 67.8\hspace{0.02em}$\pm$\hspace{0.02em}1.5          & 79.6\hspace{0.02em}$\pm$\hspace{0.02em}1.2          & 84.5\hspace{0.02em}$\pm$\hspace{0.02em}0.8          & 86.9\hspace{0.02em}$\pm$\hspace{0.02em}0.3          & 88.2\hspace{0.02em}$\pm$\hspace{0.02em}0.5          & 89.2\hspace{0.02em}$\pm$\hspace{0.02em}0.2          & 90.6\hspace{0.02em}$\pm$\hspace{0.02em}0.2          & 90.8\hspace{0.02em}$\pm$\hspace{0.02em}0.1 \\
&
\textcolor{red}{-0.6} & 
\textcolor{red}{-2.7} & 
\textcolor{red}{-2.1} & 
\textcolor{red}{-7.3} & 
\textcolor{red}{-7.9} & 
\textcolor{red}{-7.5} & 
\textcolor{red}{-5.7} &
\textcolor{red}{-5.2} &
\textcolor{red}{-5.1} &
\textcolor{red}{-4.8} &
\textcolor{red}{-4.6} & 
\textcolor{red}{-4.8}\\ 

\cmidrule(lr){1-13}
Herding   \cite{welling2009herding}  & 
11.4\hspace{0.02em}$\pm$\hspace{0.02em}0.9          & 10.8\hspace{0.02em}$\pm$\hspace{0.02em}0.5          & 11.1\hspace{0.02em}$\pm$\hspace{0.02em}0.9          & 10.6\hspace{0.02em}$\pm$\hspace{0.02em}1.1          & 
11.7\hspace{0.02em}$\pm$\hspace{0.02em}0.9          &
26.0\hspace{0.02em}$\pm$\hspace{0.02em}3.4          & 50.1\hspace{0.02em}$\pm$\hspace{0.02em}1.3          & 71.0\hspace{0.02em}$\pm$\hspace{0.02em}0.6          & 79.1\hspace{0.02em}$\pm$\hspace{0.02em}1.8          & 84.6\hspace{0.02em}$\pm$\hspace{0.02em}0.6          &  90.4\hspace{0.02em}$\pm$\hspace{0.02em}0.1          & 90.8\hspace{0.02em}$\pm$\hspace{0.02em}0.1 \\
&
\textcolor{red}{-8.8} &
\textcolor{red}{-16.5} &
\textcolor{red}{-23.7} &
\textcolor{red}{-40.4} &
\textcolor{red}{-51.8} &
\textcolor{red}{-48.1} &
\textcolor{red}{-30.0} &
\textcolor{red}{-14.2} &
\textcolor{red}{-8.9} &
\textcolor{red}{-5.2} &
\textcolor{red}{-4.2} &
\textcolor{red}{-4.8}\\ 
k-Center Greedy \cite{sener2017active} & 
12.6\hspace{0.02em}$\pm$\hspace{0.02em}1.3          & 14.3\hspace{0.02em}$\pm$\hspace{0.02em}0.8          & 16.1\hspace{0.02em}$\pm$\hspace{0.02em}1.0          & 29.6\hspace{0.02em}$\pm$\hspace{0.02em}1.6          & 41.7\hspace{0.02em}$\pm$\hspace{0.02em}3.0          & 62.0\hspace{0.02em}$\pm$\hspace{0.02em}2.0          & 73.8\hspace{0.02em}$\pm$\hspace{0.02em}1.8          & 80.2\hspace{0.02em}$\pm$\hspace{0.02em}0.7          & 83.9\hspace{0.02em}$\pm$\hspace{0.02em}0.7          & 86.6\hspace{0.02em}$\pm$\hspace{0.02em}0.5          & 90.4\hspace{0.02em}$\pm$\hspace{0.02em}0.3          & 90.8\hspace{0.02em}$\pm$\hspace{0.02em}0.1 \\ 
&
\textcolor{red}{-5.9} &
\textcolor{red}{-12.5} &
\textcolor{red}{-15.0} &
\textcolor{red}{-21.8} &
\textcolor{red}{-34.1} &
\textcolor{red}{-25.0} &
\textcolor{red}{-17.1} &
\textcolor{red}{-12.6} &
\textcolor{red}{-10.0} &
\textcolor{red}{-7.5} &
\textcolor{red}{-5.0} &
\textcolor{red}{-4.8}\\ 
\cmidrule(lr){1-13}
Forgetting   \cite{toneva2018empirical}  & 
21.8\hspace{0.02em}$\pm$\hspace{0.02em}1.6 & 
31.1\hspace{0.02em}$\pm$\hspace{0.02em}1.0 & 
35.0\hspace{0.02em}$\pm$\hspace{0.02em}1.3 & 
52.8\hspace{0.02em}$\pm$\hspace{0.02em}1.2 & 
66.4\hspace{0.02em}$\pm$\hspace{0.02em}1.3 & 
83.2\hspace{0.02em}$\pm$\hspace{0.02em}1.0 & 
88.9\hspace{0.02em}$\pm$\hspace{0.02em}0.2 & 
90.7\hspace{0.02em}$\pm$\hspace{0.02em}0.2 & 
91.0\hspace{0.02em}$\pm$\hspace{0.02em}0.4 & 
91.5\hspace{0.02em}$\pm$\hspace{0.02em}0.4 & 
90.9\hspace{0.02em}$\pm$\hspace{0.02em}0.1 & 
90.8\hspace{0.02em}$\pm$\hspace{0.02em}0.1 \\
&
\textcolor{green}{+0.4} &
\textcolor{green}{+1.3} &
\textcolor{red}{-0.2} &
\textcolor{green}{+0.7} &
\textcolor{red}{-0.6} &
\textcolor{red}{-3.4} &
\textcolor{red}{-2.8} &
\textcolor{red}{-2.8} &
\textcolor{red}{-3.1} &
\textcolor{red}{-3.1} &
\textcolor{red}{-4.4} &
\textcolor{red}{-4.8}\\ 
GraNd  \cite{paul2021deep}   & 
11.5\hspace{0.02em}$\pm$\hspace{0.02em}0.9 & 
11.9\hspace{0.02em}$\pm$\hspace{0.02em}0.8 & 
11.1\hspace{0.02em}$\pm$\hspace{0.02em}0.6 & 
10.8\hspace{0.02em}$\pm$\hspace{0.02em}1.1 & 
10.6\hspace{0.02em}$\pm$\hspace{0.02em}1.2 & 
25.4\hspace{0.02em}$\pm$\hspace{0.02em}0.9 & 
44.8\hspace{0.02em}$\pm$\hspace{0.02em}2.0 & 
67.2\hspace{0.02em}$\pm$\hspace{0.02em}2.6 & 
79.4\hspace{0.02em}$\pm$\hspace{0.02em}1.3 & 
86.3\hspace{0.02em}$\pm$\hspace{0.02em}0.3 & 
90.2\hspace{0.02em}$\pm$\hspace{0.02em}0.2 & 
90.8\hspace{0.02em}$\pm$\hspace{0.02em}0.1 \\ 
&
\textcolor{red}{-6.2} &
\textcolor{red}{-12.1} &
\textcolor{red}{-15.6} &
\textcolor{red}{-29.0} &
\textcolor{red}{-42.1} &
\textcolor{red}{-52.8} &
\textcolor{red}{-46.4} &
\textcolor{red}{-26.5} &
\textcolor{red}{-15.2} &
\textcolor{red}{-8.7} &
\textcolor{red}{-5.3} &
\textcolor{red}{-4.8}\\ 
\cmidrule(lr){1-13}
CAL   \cite{margatina2021active}      & 
21.3\hspace{0.02em}$\pm$\hspace{0.02em}1.7          & 30.8\hspace{0.02em}$\pm$\hspace{0.02em}1.0          & 36.8\hspace{0.02em}$\pm$\hspace{0.02em}1.3          & 59.9\hspace{0.02em}$\pm$\hspace{0.02em}0.8          & 71.3\hspace{0.02em}$\pm$\hspace{0.02em}1.0          & 80.0\hspace{0.02em}$\pm$\hspace{0.02em}0.2          & 83.9\hspace{0.02em}$\pm$\hspace{0.02em}0.6          & 87.1\hspace{0.02em}$\pm$\hspace{0.02em}0.3          & 89.1\hspace{0.02em}$\pm$\hspace{0.02em}0.2          & 90.6\hspace{0.02em}$\pm$\hspace{0.02em}0.2          & 91.9\hspace{0.02em}$\pm$\hspace{0.02em}0.1          & 90.8\hspace{0.02em}$\pm$\hspace{0.02em}0.1 \\
&
\textcolor{red}{-1.4} &
\textcolor{red}{-2.3} &
\textcolor{red}{-1.0} &
\textcolor{red}{-0.1} &
\textcolor{red}{-0.5} &
\textcolor{red}{-0.9} &
\textcolor{red}{-2.1} &
\textcolor{red}{-0.4} &
\textcolor{red}{-0.3} &
\textcolor{red}{-1.0} &
\textcolor{red}{-2.8} &
\textcolor{red}{-4.8}\\ 
DeepFool     \cite{ducoffe2018adversarial}    & 
17.4\hspace{0.02em}$\pm$\hspace{0.02em}0.9          & 21.6\hspace{0.02em}$\pm$\hspace{0.02em}1.4          & 25.3\hspace{0.02em}$\pm$\hspace{0.02em}1.3          & 33.6\hspace{0.02em}$\pm$\hspace{0.02em}0.4          & 43.9\hspace{0.02em}$\pm$\hspace{0.02em}3.2          & 65.5\hspace{0.02em}$\pm$\hspace{0.02em}1.6          & 77.0\hspace{0.02em}$\pm$\hspace{0.02em}1.3          &  84.5\hspace{0.02em}$\pm$\hspace{0.02em}0.5          & 86.6\hspace{0.02em}$\pm$\hspace{0.02em}0.9          & 88.7\hspace{0.02em}$\pm$\hspace{0.02em}0.5          & 90.8\hspace{0.02em}$\pm$\hspace{0.02em}0.2          & 90.8\hspace{0.02em}$\pm$\hspace{0.02em}0.1 \\ 
&
\textcolor{red}{-0.2} &
\textcolor{red}{-0.8} &
\textcolor{red}{-2.3} &
\textcolor{red}{-9.0} &
\textcolor{red}{-16.9} &
\textcolor{red}{-17.5} &
\textcolor{red}{-13.0} &
\textcolor{red}{-8.6} &
\textcolor{red}{-7.5} &
\textcolor{red}{-6.1} &
\textcolor{red}{-4.7} &
\textcolor{red}{-4.8}\\ \cmidrule(lr){1-13}
Craig     \cite{mirzasoleiman2020coresets}   & 
19.5\hspace{0.02em}$\pm$\hspace{0.02em}1.4          & 20.2\hspace{0.02em}$\pm$\hspace{0.02em}1.4          & 24.8\hspace{0.02em}$\pm$\hspace{0.02em}1.1          & 30.4\hspace{0.02em}$\pm$\hspace{0.02em}0.9          & 31.7\hspace{0.02em}$\pm$\hspace{0.02em}1.6          & 39.2\hspace{0.02em}$\pm$\hspace{0.02em}1.5          & 58.4\hspace{0.02em}$\pm$\hspace{0.02em}2.9          & 73.1\hspace{0.02em}$\pm$\hspace{0.02em}1.4          & 81.4\hspace{0.02em}$\pm$\hspace{0.02em}0.6          & 85.3\hspace{0.02em}$\pm$\hspace{0.02em}0.4          & 90.5\hspace{0.02em}$\pm$\hspace{0.02em}0.3          & 90.8\hspace{0.02em}$\pm$\hspace{0.02em}0.1 \\
&
\textcolor{red}{-3.0} &
\textcolor{red}{-6.8} &
\textcolor{red}{-6.9} &
\textcolor{red}{-14.8} &
\textcolor{red}{-28.5} &
\textcolor{red}{-40.4} &
\textcolor{red}{-30.0} &
\textcolor{red}{-17.7} &
\textcolor{red}{-11.9} &
\textcolor{red}{-8.9} &
\textcolor{red}{-5.0} &
\textcolor{red}{-4.8}\\ 
GradMatch   \cite{killamsetty2021grad}  & 
15.7\hspace{0.02em}$\pm$\hspace{0.02em}2.0          & 21.4\hspace{0.02em}$\pm$\hspace{0.02em}0.7          & 23.0\hspace{0.02em}$\pm$\hspace{0.02em}1.6          & 27.6\hspace{0.02em}$\pm$\hspace{0.02em}2.4          & 31.4\hspace{0.02em}$\pm$\hspace{0.02em}2.5          & 37.7\hspace{0.02em}$\pm$\hspace{0.02em}2.1          & 55.6\hspace{0.02em}$\pm$\hspace{0.02em}2.5          & 72.0\hspace{0.02em}$\pm$\hspace{0.02em}1.4          & 80.3\hspace{0.02em}$\pm$\hspace{0.02em}0.4          & 85.3\hspace{0.02em}$\pm$\hspace{0.02em}0.5          & 90.2\hspace{0.02em}$\pm$\hspace{0.02em}0.2          & 90.8\hspace{0.02em}$\pm$\hspace{0.02em}0.1 \\ 
&
\textcolor{red}{-1.7} &
\textcolor{red}{-4.2} &
\textcolor{red}{-7.8} &
\textcolor{red}{-19.6} &
\textcolor{red}{-30.1} &
\textcolor{red}{-42.2} &
\textcolor{red}{-31.8} &
\textcolor{red}{-18.4} &
\textcolor{red}{-12.6} &
\textcolor{red}{-7.9} &
\textcolor{red}{-3.5} &
\textcolor{red}{-4.8}\\ \cmidrule(lr){1-13}
Glister  \cite{killamsetty2021glister}  & 
14.9\hspace{0.02em}$\pm$\hspace{0.02em}2.0          & 20.9\hspace{0.02em}$\pm$\hspace{0.02em}1.5          & 24.5\hspace{0.02em}$\pm$\hspace{0.02em}1.3          & 29.0\hspace{0.02em}$\pm$\hspace{0.02em}1.9          & 31.7\hspace{0.02em}$\pm$\hspace{0.02em}2.1          & 40.2\hspace{0.02em}$\pm$\hspace{0.02em}2.3          & 57.2\hspace{0.02em}$\pm$\hspace{0.02em}1.3          & 72.2\hspace{0.02em}$\pm$\hspace{0.02em}1.3          & 80.6\hspace{0.02em}$\pm$\hspace{0.02em}0.4          & 85.4\hspace{0.02em}$\pm$\hspace{0.02em}0.5          & 90.2\hspace{0.02em}$\pm$\hspace{0.02em}0.2 & 
90.8\hspace{0.02em}$\pm$\hspace{0.02em}0.1 \\ 
&
\textcolor{red}{-4.6} &
\textcolor{red}{-6.6} &
\textcolor{red}{-8.4} &
\textcolor{red}{-21.7} &
\textcolor{red}{-34.6} &
\textcolor{red}{-44.6} &
\textcolor{red}{-33.7} &
\textcolor{red}{-20.8} &
\textcolor{red}{-13.4} &
\textcolor{red}{-9.4} &
\textcolor{red}{-5.4} &
\textcolor{red}{-4.8}\\ \cmidrule(lr){1-13}
Facility Location     \cite{iyer2021submodular}    & 20.7\hspace{0.02em}$\pm$\hspace{0.02em}1.4 & 
24.5\hspace{0.02em}$\pm$\hspace{0.02em}0.8 & 
29.5\hspace{0.02em}$\pm$\hspace{0.02em}1.2 & 
43.1\hspace{0.02em}$\pm$\hspace{0.02em}2.9 & 
51.3\hspace{0.02em}$\pm$\hspace{0.02em}1.7 & 
61.5\hspace{0.02em}$\pm$\hspace{0.02em}2.0          & 75.8\hspace{0.02em}$\pm$\hspace{0.02em}0.6          & 81.5\hspace{0.02em}$\pm$\hspace{0.02em}0.4          & 84.9\hspace{0.02em}$\pm$\hspace{0.02em}0.3          & 87.2\hspace{0.02em}$\pm$\hspace{0.02em}0.3          & 90.6\hspace{0.02em}$\pm$\hspace{0.02em}0.2          & 90.8\hspace{0.02em}$\pm$\hspace{0.02em}0.1\\ 
&
\textcolor{red}{-1.6} &
\textcolor{red}{-7.1} &
\textcolor{red}{-9.4} &
\textcolor{red}{-17.7} &
\textcolor{red}{-23.4} &
\textcolor{red}{-24.1} &
\textcolor{red}{-15.6} &
\textcolor{red}{-11.7} &
\textcolor{red}{-9.0} &
\textcolor{red}{-7.3} &
\textcolor{red}{-4.9} &
\textcolor{red}{-4.8}\\ 
GraphCut   \cite{iyer2021submodular}    &   
22.0\hspace{0.02em}$\pm$\hspace{0.02em}0.7 & 
34.0\hspace{0.02em}$\pm$\hspace{0.02em}2.2 & 
41.3\hspace{0.02em}$\pm$\hspace{0.02em}1.2 & 
66.0\hspace{0.02em}$\pm$\hspace{0.02em}1.2 & 
77.5\hspace{0.02em}$\pm$\hspace{0.02em}0.6 & 
84.6\hspace{0.02em}$\pm$\hspace{0.02em}0.5          & 87.6\hspace{0.02em}$\pm$\hspace{0.02em}0.2          & 89.8\hspace{0.02em}$\pm$\hspace{0.02em}0.4          & 87.8\hspace{0.02em}$\pm$\hspace{0.02em}0.2          & 87.2\hspace{0.02em}$\pm$\hspace{0.02em}0.1          & 90.0\hspace{0.02em}$\pm$\hspace{0.02em}0.2          & 90.8\hspace{0.02em}$\pm$\hspace{0.02em}0.1\\  
&
\textcolor{red}{-2.3} &
\textcolor{red}{-0.9} &
\textcolor{red}{-1.5} &
\textcolor{green}{+0.3} &
\textcolor{green}{+0.9} & 
\textcolor{green}{+0.6} &
\textcolor{red}{-0.2} &
\textcolor{red}{-0.8} &
\textcolor{red}{-5.4} &
\textcolor{red}{-7.2} &
\textcolor{red}{-5.4} &
\textcolor{red}{-4.8}\\ 
\bottomrule
\end{tabular}
}
}
\end{table*}

\section{Applications}
\label{sec:applications}
Coreset selection has found applications in various domains, spanning computer vision \cite{lee2024coreset, zhou2022probabilistic, mahawar2024label, jia2024unsupervised, laribi2024application}, natural language processing \cite{attendu2023nlu, joaquin2024in2core, zhangstaff, tsai2024code}, and quantum machine learning \cite{qu2022performance, huang2024coreset, xue2023near}. 
In the following sections, we examine where coreset selection has made interesting contributions.

Across applications, the key difference is which \emph{signal} is available for selection: training-free pipelines operate in a fixed embedding space (often from foundation models) to optimize coverage/diversity, training-oriented methods leverage partial training dynamics (loss/gradients/margins) for model-specific utility, and label-free variants replace ground-truth labels with representation proxies or pseudo-labels (scalable but brittle under domain shift).

\subsection{Image Enhancement}
Coreset selection has also proven valuable in image enhancement tasks, specifically in the domain of image super-resolution, where practitioners often construct training sets by randomly cropping patches from high-resolution images (800 2K images lead to 32K training images) \cite{moser2024study, ding2023not, laribi2024application}. 
Although simple, this patch-based extraction frequently yields a large proportion of redundant low-frequency patches that convey limited additional information \cite{kong2021classsr, moser2023yoda}. 

By adopting a score-based training-oriented coreset selection strategy, one that ranks patches based on their importance to the model’s learning dynamics (i.e., Sobel filter or loss value-based), it becomes possible to discard uninformative patches in favor of fewer but more critical (high-frequency) examples. 
This not only leads to a substantial reduction in dataset size and training overhead, but recent empirical results further show that such curated training sets can improve the reconstruction quality of super-resolution networks. 

In addition, we can use score-based patch ranking to weight image patches for deeper refinement during training, as proposed by ClassSR \cite{wang2022adaptive} and others \cite{moser2023dynamic, xiao2021self, kong2024towards, luo2024skipdiff}.

\subsection{Neural Architecture Search}
Neural Architecture Search (NAS) automates the design of network topologies and sometimes outperforms manually engineered architectures at the cost of substantial computational overhead \cite{real2019regularized, pham2018efficient}. 
A key bottleneck arises when evaluating candidate architectures on large datasets, as each candidate in the search space must be trained (or partially trained) to assess its fitness \cite{liu2018darts, dong2019searching}. 
This computational burden grows significantly when the dataset itself is extensive \cite{liu2022survey, kang2023neural}. 

Recent work shows that coreset selection can overcome these scalability issues by constructing coresets, i.e., smaller yet sufficiently representative subsets used during the architecture search phase \cite{na2021accelerating, moser2022less, yao2023asp, zhou2020econas}. 
For instance, ranking and filtering out less informative training samples often shortens the NAS procedure by a factor of two or more while simultaneously improving the resulting architecture’s accuracy once retrained on the full dataset. 

\subsection{Dataset Distillation}
Coreset selection and dataset distillation share the same high-level goal: reducing the size of a dataset. 
Yet, they differ in the nature of the compressed data. 
In dataset distillation, one synthesizes entirely new samples that preserve training quality \cite{cazenavette2023generalizing, moser2024latent, liu2025evolution}. 

Concretely, let the original image classification dataset be $\mathcal{T} = (X_r, Y_r)$ with $N$ real examples. 
Dataset distillation tries to generate a small, synthetic set $\mathcal{S} = (X_s, Y_s)$ of $M \ll N$ synthetic samples. 
We define this objective as
\begin{equation}
    \mathcal{S}^* = \arg\min_{\mathcal{S}} \mathcal{L}\bigl(\mathcal{S}, \mathcal{T}\bigr),
\end{equation}
where $\mathcal{L}$ measures how good the synthetic set $\mathcal{S}$ represents the original dataset $\mathcal{T}$. 
Concrete implementations of $\mathcal{L}$ are, for example, methods like Dataset Condensation \cite{zhao2020dataset}, Distribution Matching \cite{zhao2023dataset}, or Matching Training Trajectories \cite{cazenavette2022dataset} and others \cite{yin2023squeeze, su2024d}.

Since both coreset selection and dataset distillation ultimately aim for the same destination, it is only natural to explore how coreset and distillation methods might be combined for even more powerful dataset reduction \cite{moser2024distill, chenunified, khandel2024distillation}.

For instance, \textit{He et al.} \cite{he2023yoco} apply coreset selection directly to the distilled dataset by combining a logit-based prediction-error criterion (aiming to retain ``easy'' samples) with a Rademacher-complexity measure that maintains class balance. 
Separately, \textit{Xu et al.} \cite{xu2024distill} propose a complementary pipeline in which coreset selection, guided by empirical loss, prunes real data before the distillation stage and is further enhanced by a dynamic pruning mechanism during the distillation process.

\subsection{Imbalanced Datasets}
Class imbalance poses significant challenges for machine learning models, often leading to biased predictions toward majority classes \cite{zhang2023fed, soremekun2022astraea}. 
Coreset selection techniques have been explored as a potential solution to this problem \cite{aggarwal2021minority}.

For instance, in federated learning, where multiple clients collaborate in training together, class imbalance is a major issue. 
It has been shown that adapting the principles of coreset selection to this scenario can mitigate the imbalance problem \cite{sivasubramanian2024gradient, luo2024dual}.
As a result, coordinating coreset selection across clients achieves a globally balanced dataset while respecting the privacy constraints necessary in federated settings.

\subsection{Continual Learning}
Continual Learning (CL) aims to enable machine learning models to learn new tasks while preserving previously acquired knowledge incrementally \cite{thrun1995lifelong, wang2024comprehensive}. 
A fundamental challenge in CL is catastrophic forgetting, where a model forgets prior knowledge when trained on new data \cite{mccloskey1989catastrophic}. 
For instance, a model could encounter classes in the current tasks that are not appearing in later tasks, which results in forgetting them.
Rehearsal-based strategies, which maintain a subset of past examples for replay during training, have emerged as an effective solution to mitigate forgetting \cite{titsias2019functional, mirzadeh2020understanding}. 

However, storing and retraining on large amounts of past data is computationally expensive and memory-intensive. 
Coreset selection techniques play a crucial role in optimizing the selection of representative samples for rehearsal, ensuring efficient learning while reducing memory overhead \cite{nguyen2017variational, borsos2020coresets, yoon2021online}.

\subsection{Reinforcement Learning}
Similar to CL, Reinforcement Learning (RL) requires extensive memory sampling to not forget past experiences \cite{hafez2023map}.
As RL is naturally operating in interactive environments (e.g., in robotics), retraining agents also in parallel on past experiences is costly in terms of both data and computation \cite{sutton2018reinforcement}. 
Thus, coreset selection has been used to improve experience replay by selecting the most informative past experiences for training \cite{zhan2025coreset, zheng2024selective}.

\subsection{Machine Unlearning}
Machine unlearning refers to the process of removing the influence of specific data points from a trained machine learning model without retraining it. This is particularly useful in scenarios where user privacy concerns require the deletion of data, or when models must ``forget" outdated or incorrect information. 

Coreset selection can play a crucial role in this process by identifying the most important data points to either retain or remove, thereby minimizing the impact on the model's overall performance while improving the cost of unlearning \cite{machine-unlearning-1912-03817}. Utility Preserving Coreset Selection (UPCORE) is a framework introduced by \cite{patil2025} for data selection that minimizes the performance degradation during unlearning, focusing on which data points to remove (i.e., the "forget set") but also leveraging a coreset selection algorithm to identify which samples to retain. 

\subsection{Self-supervised Learning}
Using coresets in self-supervised learning can improve the efficiency and effectiveness of SSL pre-training. This is critical when training on large open sets (e.g., ImageNet) for domain-specific tasks (e.g., fine-grained classification) where distribution mismatch between the open set and target dataset can degrade performance. \textit{Kim et al.} \cite{kim2023coreset} propose Simcore, a technique that samples a coreset that minimizes the latent-space distance between the open-set and target dataset. \textit{Singh et al.} \cite{singh2024bloomcoreset} utilize Bloom filters for rapid retrieval of the coreset from the open set and integrate it with Simcore to achieve a reduction in sampling time. RETRIEVE \cite{killamsetty2021retrieve} achieves a speedup of around 3$\times$ in the traditional SSL setting. 

Another emerging direction is to increasingly use \emph{foundation-model} embedding spaces, often trained self-supervised, as the default geometry for training-free and label-free selection, e.g., DINO \cite{caron2021emerging}, or DINOv2 \cite{oquab2023dinov2} representations, which tend to be more semantically aligned than task-trained backbones.

\subsection{LLM Pretraining}
LLM pretraining corpora span trillions of tokens, making coreset selection a practical necessity: RedPajama \cite{weber2024redpajama} approximates the LLaMA pipeline by removing low-quality Common Crawl documents via score-based signals (e.g., CCNet perplexity sorting to drop the high-perplexity tail), while Dolma \cite{soldaini2024dolma} curates roughly 200TB of raw text into a cleaned 11TB (about 3T tokens) dataset through label-free quality filters (e.g., boilerplate stripping and removal of nonsensical/offensive text). Beyond quality filtering, training-free geometry-based selection is central for redundancy reduction at web scale: SlimPajama \cite{he2024softdedup} applies global MinHash-LSH soft deduplication to remove near-duplicate passages, eliminating about 49.6\% of 1.2T tokens and yielding a 627B-token subset with higher information density, while corpus mixing across modalities/sources (code, books, web, scientific text) in RedPajama and Dolma can be interpreted as coarse cluster coverage for diversity. Training-based scoring further aligns selection with downstream learning signals: RefinedWeb \cite{penedo2023refinedweb} demonstrates that aggressive filtering and deduplication of Common Crawl using learned criteria (including “Gopher”-style rules to discard gibberish and overly simplistic content) produces a high-quality web corpus (with a released 600B-token extract) that improves model outcomes, and large-scale dataset construction efforts such as The Pile \cite{gao2020pile} and ROOTS \cite{laurenccon2022bigscience} emphasize broad domain coverage (implicitly echoing submodular/coverage objectives) even when explicit submodular optimization is not directly applied at trillion-token scales.

\section{Discussion and Future Directions}
\label{sec:discussion}
Coreset selection has evolved significantly beyond its early geometric origins, fueled by deep learning’s growing scale and resource needs. 
As a result, coreset selection finds itself applied to many machine learning tasks.
As surveyed, training-free methods remain conceptually straightforward and computationally light, yet often underperform in comparison to training-oriented methods in tasks demanding fine-grained or model-specific subsets. 

On the other hand, training-based approaches, particularly those relying on loss signals, gradient information, and submodular formulations, can yield competitive or even state-of-the-art performance, albeit at a higher computational cost. 
The recent development in active learning and label-free coreset selection further broadens the field, approaching fully label-free environments.

\subsection{Discussion}
Despite these advancements, we observe the following:

\textbf{Trade-offs Between Simplicity and Specificity.} Training-free methods (e.g., random, Herding, and k-Center Greedy) perform surprisingly well in some scenarios and scale effortlessly to massive datasets. 
By contrast, training-oriented or bilevel approaches (e.g., GradMatch, GLISTER) finely tune subsets for a specific architecture or objective but can be sensitive to random seeds, hyperparameters, and model changes.

\textbf{Coverage vs. Difficulty.}
A recurring practical question is: \emph{How do we keep decision-boundary information \emph{and} preserve distributional coverage under aggressive pruning?}
Hard-sample scoring (GraNd/EL2N, boundary-based) tends to sharpen boundaries but can collapse coverage and over-select outliers, especially at high pruning ratios.
Conversely, coverage-centric families (submodular facility-location/graph-cut, $k$-Center) preserve modes but may miss boundary refinement.
This suggests hybrids that explicitly mix \emph{coverage} (training-free/submodular) with \emph{difficulty} (training-oriented scores), potentially with a budget split that adapts with dataset size and pruning ratio, similar to \textit{Zheng et al.} \cite{zheng2022coverage}.

\textbf{Label-Free Approaches.}
In many real settings, the question is: \emph{Can we select a useful subset \emph{without} labels while remaining robust to distribution shift?}
Purely label-free pipelines like ELFS and ZCore (clustering/pseudo-labeling or VLM-based embeddings + coverage) scale well, but their failure mode is brittle representations: pseudo-label noise and embedding misalignment can dominate the subset.
In this regime, coverage/submodular selection in transferable embeddings is typically safer than pseudo-label-driven “hardness” scoring, and robustness can be improved by coupling selection with simple OOD/outlier filters or agreement signals (e.g., multi-view consistency).

\textbf{Influence of Dataset Size and Pruning Ratio.}

A practical selection question is: \emph{Which family should we use when $|\mathcal{T}|$ and $(1-\alpha)$ change?}
Empirically, at extreme compression (very small $(1-\alpha)$), coverage-oriented selectors and submodular objectives are often preferable to avoid mode dropping, whereas at moderate compression, training-oriented scoring can better target model-specific utility.
This also interacts with noise: hardest-first criteria can become outlier-dominated when labels are corrupted, suggesting either robustified scoring (smoothing/clipping) or coverage-first selection when noise is suspected.

\textbf{Evaluation and Benchmarking.}
A key open question is: \emph{Which coreset families transfer across architectures and remain reliable when the downstream model or domain changes?}
Training-oriented selectors can be tightly coupled to the scoring model and training recipe, while embedding-based training-free/submodular selectors may transfer better if the representation is strong.
This calls for benchmarks that explicitly vary (i) architecture, (ii) fine-tuning protocol, and (iii) domain shift/noise, and report both \emph{selection cost} and \emph{downstream utility} rather than accuracy alone.

\textbf{Summary.}
For high pruning ratios, submodular function maximization seems a good strategy, effectively selecting a diverse and representative subset of the data. 
However, for low pruning ratios, score-based methods are preferable due to their efficiency in retaining the most informative samples without excessive computational overhead (e.g., forgetting). 
That said, submodular approaches must be used with caution, as they can become computationally intractable if the resulting coreset remains large, even at high pruning ratios. 
Balancing selection quality with computational feasibility remains a key challenge in practical applications.

\subsection{Future Research Directions}
We summarize future directions as concrete research questions and typical deployment scenarios, and indicate which method families appear most suitable in each case.

\textbf{Adaptive and Continual Coresets.}
How can we maintain a small replay buffer that remains representative as the stream distribution drifts?
This scenario favors \emph{online coverage} (incremental $k$-Center/submodular updates in an evolving embedding space) combined with \emph{lightweight training signals} (periodic scoring refresh, forgetting-style dynamics) to prevent replay collapse onto stale modes.
Bilevel objectives are attractive when a small validation buffer exists per phase, but their cost motivates approximate, amortized updates.

\textbf{Fairness and Robustness.}
Can we design selectors that avoid outlier domination and remain stable under corrupted labels or adversarial contamination?
Submodular coverage/diversity is a strong baseline because it spreads budget across modes, while bilevel selection can explicitly target robust validation objectives when trustworthy validation data exists.
A promising direction is to integrate \emph{robust scoring} (clipped losses, uncertainty-aware weighting) with \emph{explicit OOD filtering} prior to selection, and to evaluate selectors by worst-group or worst-slice performance, not only mean accuracy.

\textbf{Self-Evolving Coresets with Meta-Learning.}
Can we meta-learn selection policies that generalize across datasets, model families, and training recipes?
This naturally aligns with (i) amortized scoring models that predict utility from representations, and (ii) bilevel/meta-learning formulations where the outer objective measures transfer to new tasks.
A key challenge is avoiding overfitting to a single benchmark: future work should evaluate cross-architecture transfer and robustness to changes in optimizer, augmentation, and label quality. 

\textbf{Multi-Modality.}
For VLM/LLM fine-tuning, what subset best preserves coverage over domains, styles, and rare capabilities while controlling redundancy?
Here, training-free selection in joint embedding spaces (CLIP-like) and submodular objectives over multimodal similarity are natural first choices, because they directly target diversity/coverage without requiring task labels.
When downstream preferences are known (e.g., instruction-following, safety), bilevel/validation-driven selection becomes compelling, using held-out task metrics as the outer objective.

\section{Conclusion}
\label{sec:conclusion}
This survey has traced the evolution of coreset selection from its geometric foundations to more sophisticated training-oriented and label-free methods. By examining core approaches - random and geometry-based sampling, gradient matching, submodular formulations, and bilevel optimization - this work reveals how different strategies strike varied balances between computational efficiency, label requirements, and accuracy. 

From a bird’s-eye view, we highlight how both small and large datasets can be pruned effectively, though the preferred criteria may shift as sample sizes grow or task complexities change. 
Furthermore, emerging perspectives on coverage, margin-based difficulty, and meta-learning provide broader frameworks for understanding what genuinely constitutes the most “informative” data. 

Taken together, these developments underscore how data pruning is not simply about efficiency but can boost generalization by preserving the crucial patterns of the original dataset. 
As coreset selection continues to mature, it stands as a promising research direction to mitigate the resource demands of modern deep learning, adapt to new domains where little to no supervision is available, and potentially unify competing priorities such as fairness, robustness, and interpretability.

\bibliography{mybibfile}

\newpage

\begin{IEEEbiography}[{\includegraphics[width=1in,height=1in,clip,keepaspectratio]{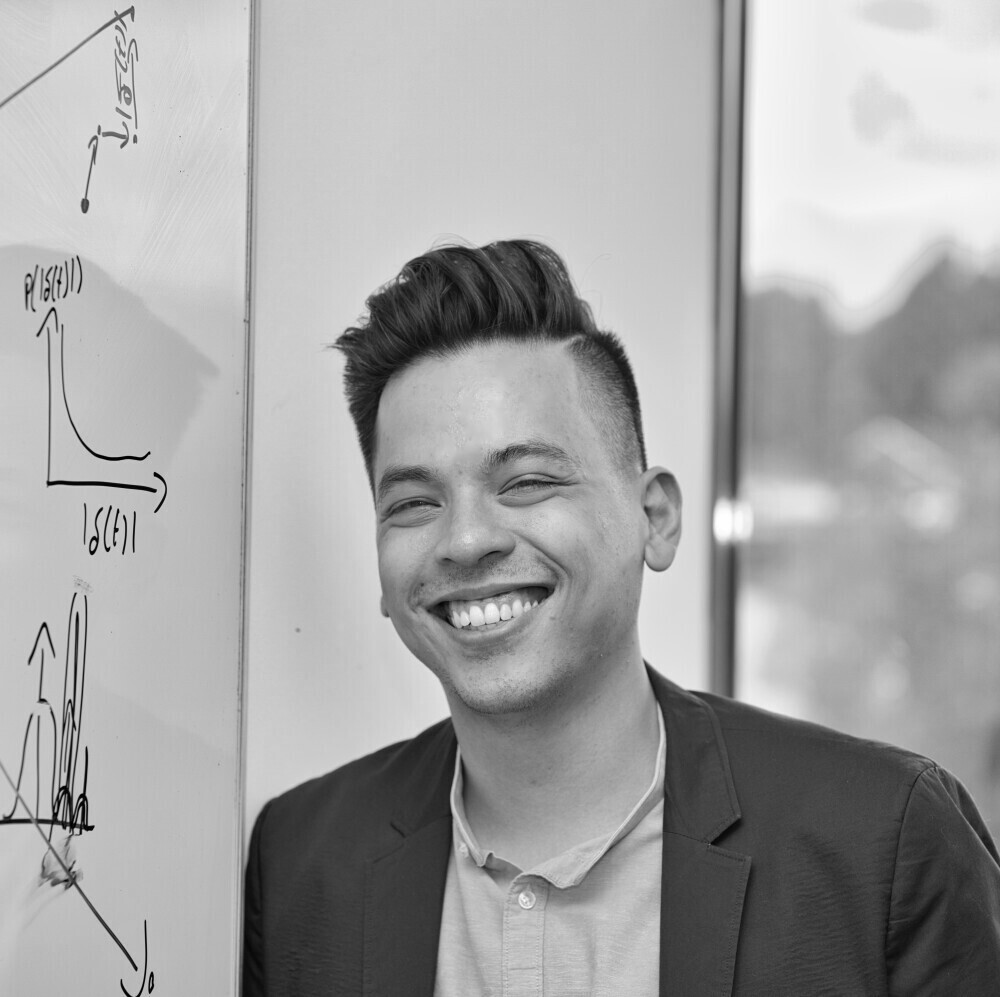}}]{Brian B. Moser}
 is a Ph.D. student at the University of Kaiserslautern-Landau and research assistant at the German Research Center for Artificial Intelligence (DFKI) in Kaiserslautern. He received the M.Sc. degree in computer science from the TU Kaiserslautern in 2021. His research interests include image super-resolution and deep learning.
\end{IEEEbiography}

\vskip -4.1\baselineskip plus -1fil

\begin{IEEEbiography}[{\includegraphics[width=1in,height=1in,clip,keepaspectratio]{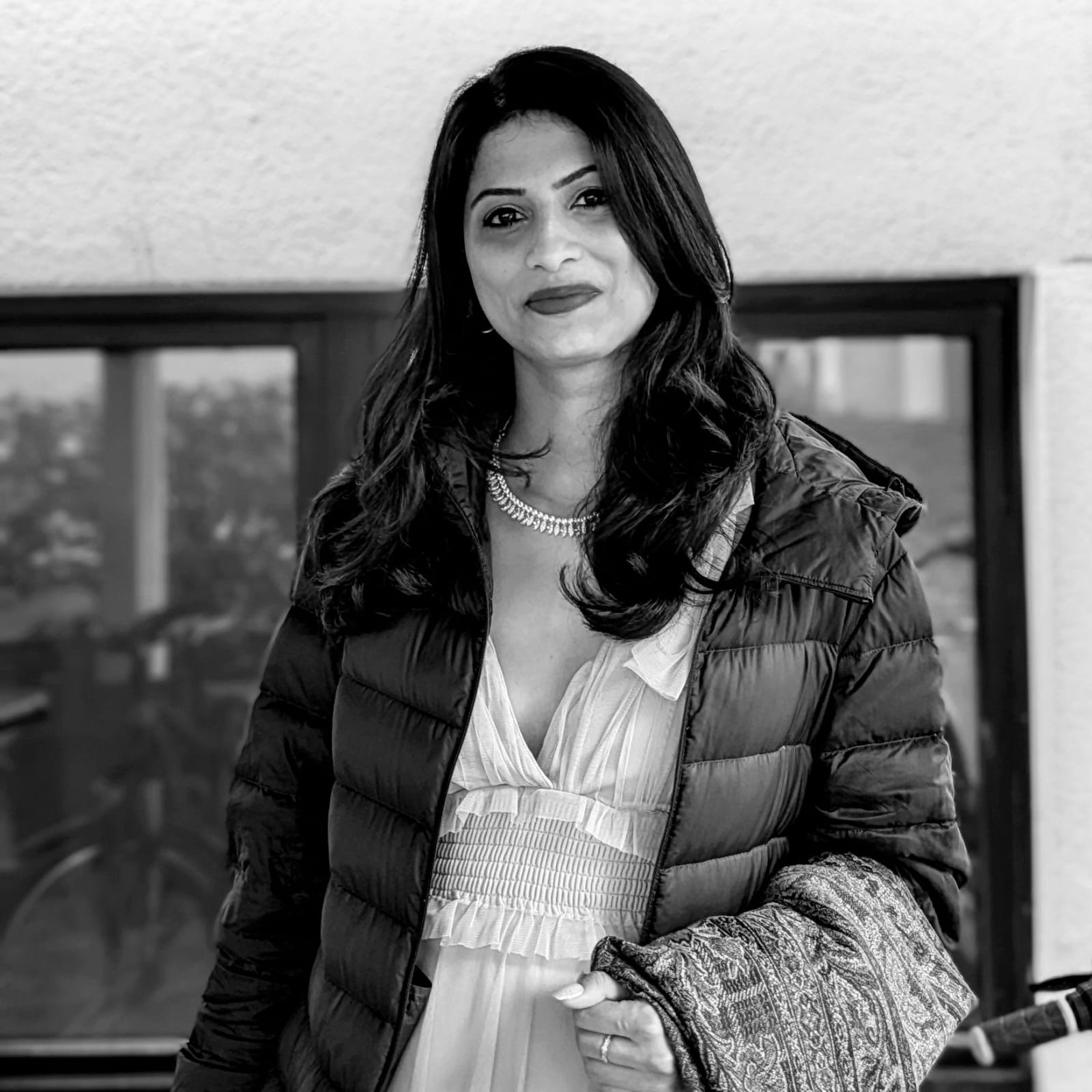}}]{Arundhati S. Shanbhag}
is a Ph.D. student at the University of Kaiserslautern-Landau and research assistant at the German Research Center for Artificial Intelligence (DFKI) in Kaiserslautern. Her research interests include computer vision and deep learning. 
\end{IEEEbiography}

\vskip -4.1\baselineskip plus -1fil

\begin{IEEEbiography}[{\includegraphics[width=1in,height=1in,clip,keepaspectratio]{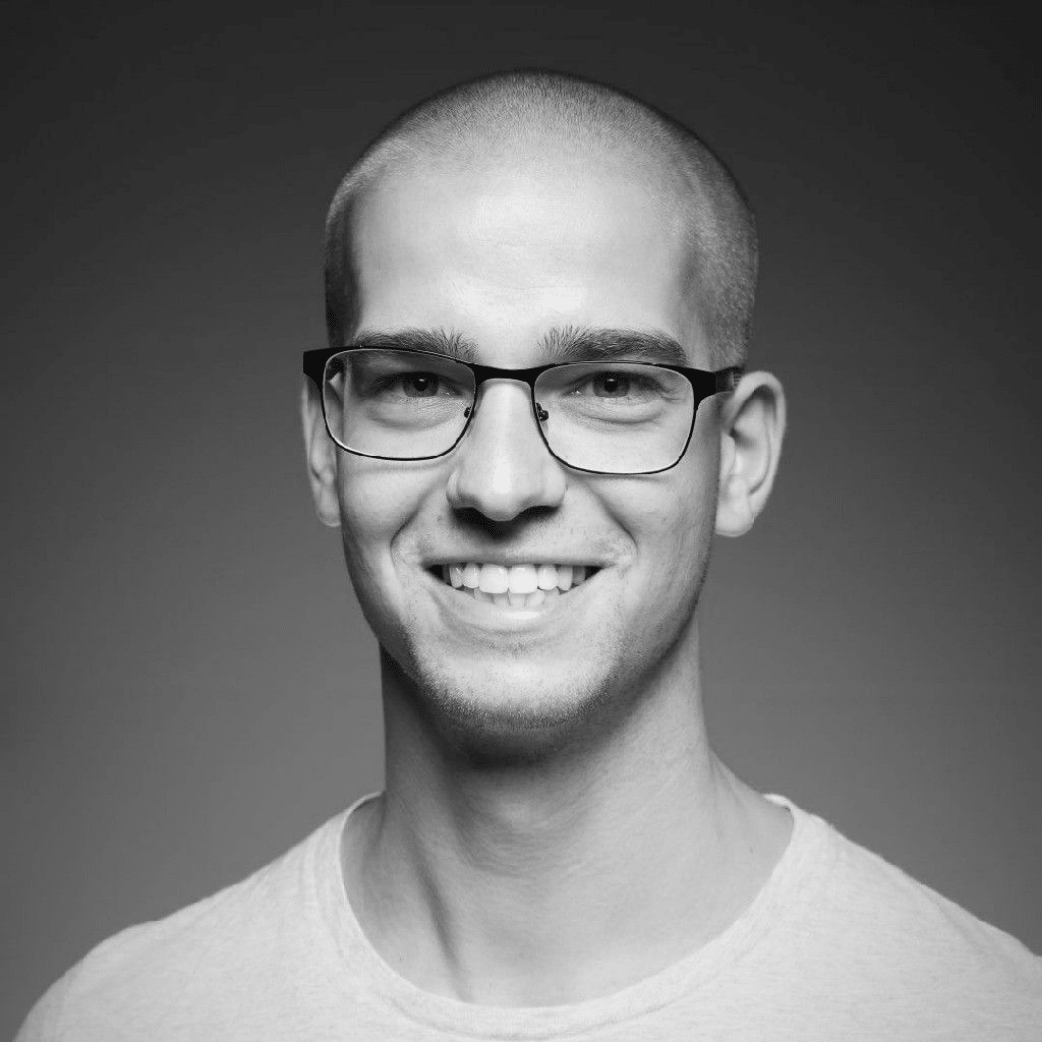}}]{Stanislav Frolov}
is a Ph.D. student at the University of Kaiserslautern-Landau and research assistant at the German Research Center for Artificial Intelligence (DFKI) in Kaiserslautern. He received the M.Sc. degree in electrical engineering from the Karlsruhe Institute of Technology in 2017. His research interests include generative models and deep learning.
\end{IEEEbiography}

\vskip -4.1\baselineskip plus -1fil

\begin{IEEEbiography}[{\includegraphics[width=1in,height=1in,clip,keepaspectratio]{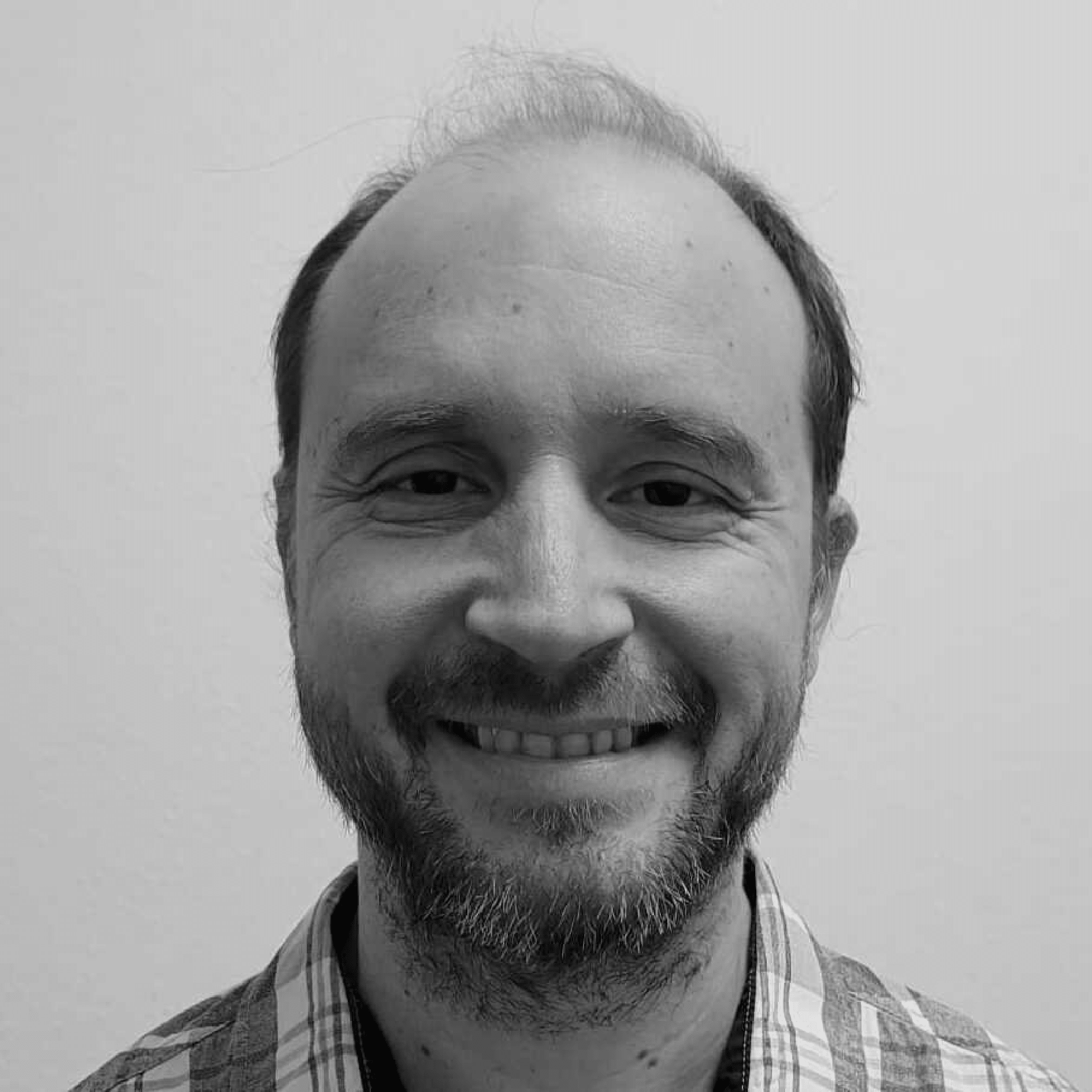}}]{Federico Raue}
 is a Senior Researcher at the German Research Center for Artificial Intelligence (DFKI) in Kaiserslautern. He received his PhD. degree at TU Kaiserslautern in 2018 and his M.Sc.  degree in Artificial Intelligence from Katholieke Universiteit Leuven in 2005. His research interests include meta-learning and multimodal machine learning.
\end{IEEEbiography}

\vskip -4.1\baselineskip plus -1fil

\begin{IEEEbiography}[{\includegraphics[width=1in,height=1in,clip,keepaspectratio]{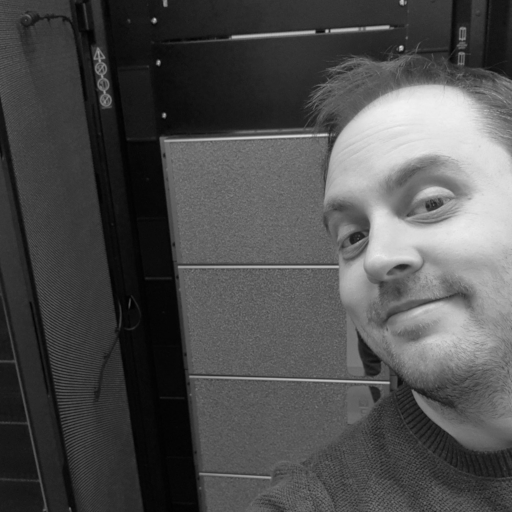}}]{Joachim Folz}
is a Senior Consultant at the German Research Center for Artificial Intelligence (DFKI) in Kaiserslautern. He received his M.Sc. degree in Computer Science with a specialization in Intelligent Systems from TU Kaiserslautern in 2014. His research interests include efficient neural networks and alternative learning methods beyond backpropagation.
\end{IEEEbiography}

\vskip -4.1\baselineskip plus -1fil

\begin{IEEEbiography}[{\includegraphics[width=1in,height=1in,clip,keepaspectratio]{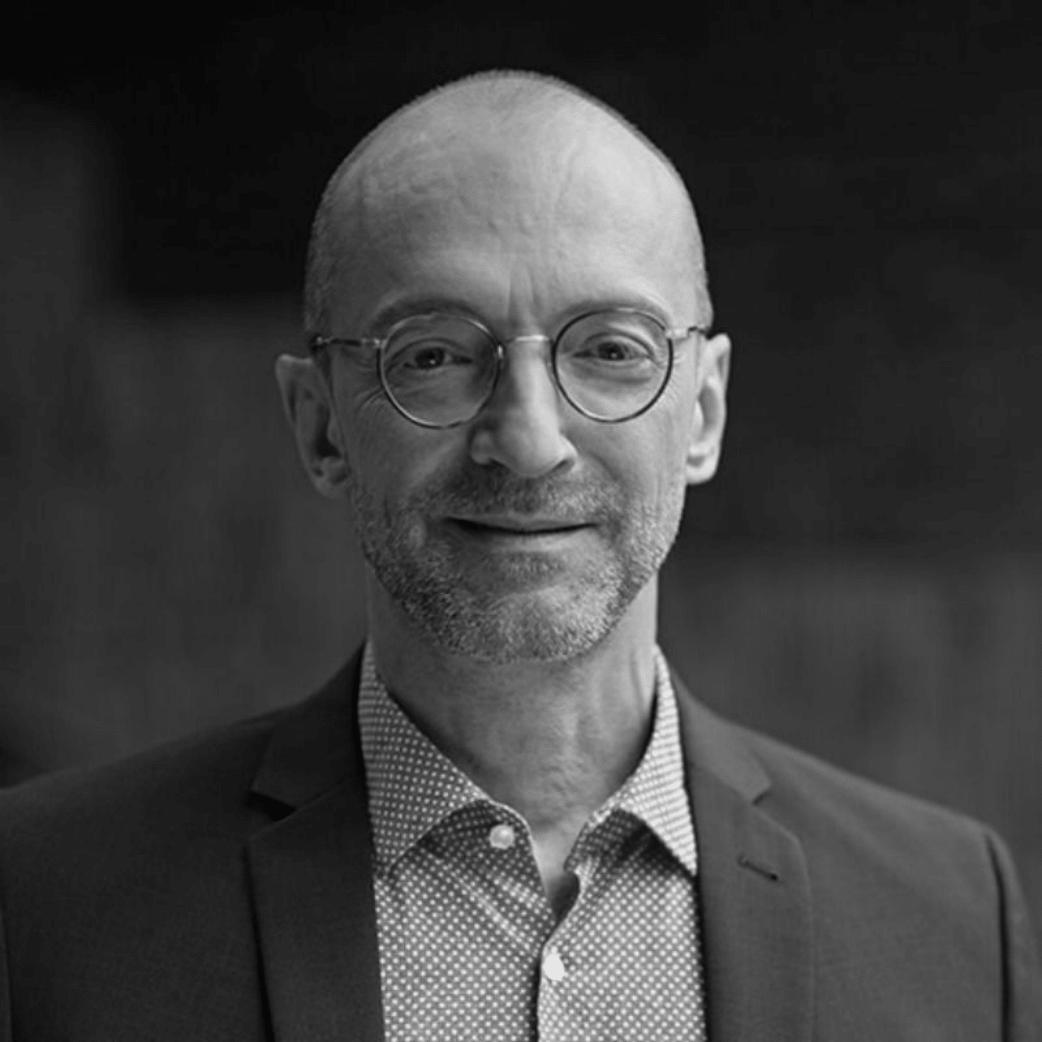}}]{Andreas Dengel}
is a Professor at the Department of Computer Science at TU Kaiserslautern and Executive Director of the German Research Center for Artificial Intelligence (DFKI) in Kaiserslautern, Head of the Smart Data and Knowledge  Services research area at DFKI and of the DFKI Deep Learning Competence Center. His research focuses on machine learning, pattern recognition, quantified learning, data mining, semantic technologies and document analysis.
\end{IEEEbiography}

\end{document}